\documentclass[10pt,twocolumn,letterpaper]{article}

\usepackage[pagenumbers]{cvpr} 
\usepackage[dvipsnames]{xcolor}
\newcommand{\red}[1]{{\color{red}#1}}

\definecolor{cvprblue}{rgb}{0.21,0.49,0.74}
\usepackage[pagebackref,breaklinks,colorlinks,citecolor=cvprblue]{hyperref}

\usepackage{blindtext}
\usepackage{graphicx}
\usepackage{caption}
\usepackage{bm}

\usepackage{multirow}
\usepackage{amssymb}
\usepackage{pifont}
\usepackage{array}
\usepackage{colortbl}
\usepackage[flushleft]{threeparttable}
\usepackage{comment}
\usepackage{float}
\usepackage{balance}
\usepackage[normalem]{ulem}
\usepackage{array}
\usepackage[T1]{fontenc}

\def\confName{CVPR\xspace}
\def\confYear{2024\xspace}
\def\modelName{AMUSE\xspace}

\title{Emotional Speech-driven 3D Body Animation via Disentangled Latent Diffusion}
\author{Kiran Chhatre\textsuperscript{1}\qquad
Radek Daněček\textsuperscript{2}\qquad
Nikos Athanasiou\textsuperscript{2}\\
Giorgio Becherini\textsuperscript{2}\qquad
Christopher Peters\textsuperscript{1}\qquad
Michael J. Black\textsuperscript{2}\qquad
Timo Bolkart\textsuperscript{2}\thanks{Now at Google.}\\
\textsuperscript{1}KTH Royal Institute of Technology, Sweden
\textsuperscript{2}Max Planck Institute for Intelligent Systems, Germany
}

\begin{document}
\maketitle
\newcommand{\rebut}[1]{\noindent\textcolor{black}{{#1}}\xspace}
\renewcommand{\paragraph}[1]{\noindent\textbf{#1}}
\newcommand{\todo}[1]{\PackageWarning{}{Unprocessed todo}
{\footnotesize \textcolor{red}{XXX \textbf{#1} XXX}}}

\newcommand{\mask}{S}
\newcommand{\mesh}{M}
\newcommand{\shape}{\bm{\beta}}
\newcommand{\pose}{\bm{\theta}}
\newcommand{\expression}{\bm{\psi}}
\newcommand{\cam}{C}
\newcommand{\imgRGB}{I}

\newcommand{\smplx}{\mbox{SMPL-X}\xspace}
\newcommand{\smplX}{\smplx}

\newcommand{\twoD}{2D\xspace}
\newcommand{\threeD}{3D\xspace}
\newcommand{\sixD}{6D\xspace}

\newcommand{\audionet}{\tau}
\newcommand{\tcnae}{\nu}
\newcommand{\enccontent}{E_{c}}
\newcommand{\encemotion}{E_{e}}
\newcommand{\encstyle}{E_{s}}
\newcommand{\dec}{D}
\newcommand{\ser}{S}
\newcommand{\mfcc}{a}
\newcommand{\latentdim}{256}
\newcommand{\latentdimstyle}{128}

\newcommand{\tmplatentcont}{c}
\newcommand{\tmplatentstyle}{s}
\newcommand{\tmplatentemo}{e}
\newcommand{\tmplatentraw}{z_r}
\newcommand{\tmpenccontent}{\mathcal{E}_c}
\newcommand{\tmpencstyle}{\mathcal{E}_s}
\newcommand{\tmpdeccontent}{\mathcal{D}_c}
\newcommand{\tmpdecstyle}{\mathcal{D}_s}
\newcommand{\tcnaudionet}{\nu_s, \nu_c}

\newcommand{\tmp}{^{1:T}}
\newcommand{\tmpfull}{^{1:L}}
\newcommand{\supmat}{Sup.~Mat.\xspace}

\newcommand{\joints}{J}
\newcommand{\numjoints}{47}
\newcommand{\priorenc}{\mathcal{P}_{E}}
\newcommand{\priordec}{\mathcal{P}_{D}}

\newcommand{\normalvec}{z_n}
\newcommand{\gaussian}{\mathcal{N}(\textbf{0},\textbf{I})}
\newcommand{\denoisernet}{\Delta }
\newcommand{\latentmotion}{z_m}
\newcommand{\priordecodemotion}{\hat{m}^{1:T}}
\newcommand{\diffdecodemotion}{\tilde{m}^{1:T}}
\newcommand{\denoisedlat}{z_{\tilde{m}}}
\newcommand{\diffdecodemotionlatent}{z_{\tilde{m}^{1:T}}}
\newcommand{\latentmotiondim}{d_z}
\newcommand{\motion}{m}
\newcommand{\vertices}{V}
\newcommand{\motiondim}{d_m}
\newcommand{\motiondist}{\textbf{M}}

\newcommand{\mfeatxtractor}{M}

\newcommand{\cmark}{\ding{51}}%
\newcommand{\xmark}{\ding{55}}%
\newcommand{\denoiser}{\Delta}
\newcommand{\denoiserpred}{\delta}
\newcommand{\pe}[1]{\textup{SE}(#1)}
\newcommand{\difft}{t_d}
\newcommand{\sg}[1]{\textup{sg}\left[#1\right]}

\newcommand{\smoothL}{L_1^s}

\newcommand{\losscontent}{\mathcal{L}_{con}}
\newcommand{\lossemotion}{\mathcal{L}_{emo}}
\newcommand{\lossstyle}{\mathcal{L}_{sty}}
\newcommand{\lossselfrec}{\mathcal{L}_{self}}
\newcommand{\losscrosscon}{\mathcal{L}_{xcon}}
\newcommand{\losscrossemo}{\mathcal{L}_{xemo}}
\newcommand{\losscrosssty}{\mathcal{L}_{xsty}}

\begin{abstract}

Existing methods for synthesizing 3D human gestures from speech have shown promising results, but they do not explicitly model the impact of emotions on the generated gestures.
Instead, these methods directly output animations from speech without control over the expressed emotion.
To address this limitation, we present \modelName 
, an emotional speech-driven body animation model based on latent diffusion. 
Our observation is that content (i.e., gestures related to speech rhythm and word utterances), emotion, and personal style are separable. 
To account for this, \modelName maps the driving audio to three disentangled latent vectors: one for content, one for emotion, and one for personal style. 
A latent diffusion model, trained to generate gesture motion sequences, is then conditioned on these latent vectors. 
Once trained, \modelName synthesizes 3D human gestures directly from speech with control over the expressed emotions and style by combining the content from the driving speech with the emotion and style of another speech sequence.
Randomly sampling the noise of the diffusion model further generates variations of the gesture with the same emotional expressivity.
Qualitative, quantitative, and perceptual evaluations demonstrate that \modelName outputs realistic gesture sequences.
Compared to the state of the art, the generated gestures are better synchronized with the speech content, and better represent the emotion expressed by the input speech.
Our code is available at \href{https://amuse.is.tue.mpg.de/}{amuse.is.tue.mpg.de}.
   
\end{abstract}
  
\section{Introduction}
\label{sec:intro}

\begin{figure}[t!]
  \centering
  \includegraphics[width=1\linewidth]{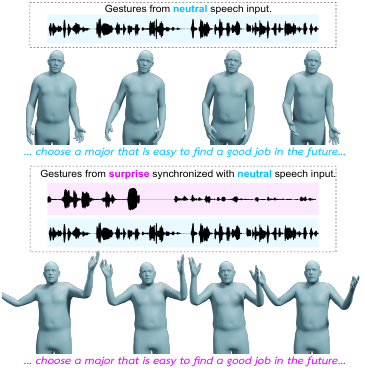}

   \captionof{figure}{\textbf{Goal.} 
   \modelName generates realistic emotional 3D body gestures directly from a speech sequence (top). 
   It provides user control over the generated emotion by combining the driving speech sequence with a different emotional audio (bottom).}
    \label{fig:teaser}
\end{figure}

Animating 3D bodies from speech has a wide range of applications, such as telepresence in AR/VR, avatar animation in games and movies, and to embody interactive digital assistants.
While methods for speech-driven 3D body animation have recently shown great progress \cite{yi2023generating,3dconvgesture_2021,Ao2023GestureDiffuCLIP,alexanderson2023listen,liu2024emage}, existing methods do not adequately address one crucial factor: the impact of emotion \rebut{from the driving speech signal} on the generated gestures.
Emotions and their expressions play a fundamental role in human communication \cite{edward2005,1992mcneil,Herbert2012GestureTF} and have become an important consideration when designing computer systems that interact with humans in a natural manner \cite{PicardAffective1997,  HUMAINE2011}. They are of central concern when synthesizing human animations for a wide variety of application contexts, such as Socially Interactive Agents \cite{SIAs2021}. Because of this, speech-driven animation systems must not only align movement with the rhythm of the speech, but should also be capable of generating gestures that  are perceived as expressing the suitable emotion.

Many factors contribute to the perception of emotion and personal idiosyncrasies, such as facial expressions \cite{ekman1993}, gaze and eye contact \cite{kleinke1986}, physiological responses \cite{levenson2003}, tone of voice \cite{scherer2003}, body language \cite{mehrabian1972}, and gestures \cite{kendon2004}. 
When it comes to 3D animation, the most relevant factors are facial expressions, gestures, and body language~\cite{williams2012animator}.
While emotional speech-driven animation methods have recently been proposed for 3D faces \cite{peng2023emotalk, danecek2023emotional,laughtalk,zhong2023expclip}, animating emotional bodies from speech remains under-explored. 

Generating gestures solely from speech with emotional control is a difficult task. 
First, the mapping from audio to body motion is a non-deterministic many-to-many mapping, which is difficult to model.
Gestures across subjects can vary when uttering the same sentence, and a single individual's motions can change significantly across repetitions. 
Second, factoring out the impact of emotional state on the body motion from other, unknown factors, is difficult.
This requires disentangling the effects of three different factors on the generated motion, namely content-based (i.e., gestures related to speech rhythm and word utterances), emotion-based, and those based on personal style. 
\modelName addresses this by separating a speech sequence into content, emotion, and style latent vectors, which are then used to condition a latent diffusion model. 
Specifically, \modelName consists of three main components:
(1) an audio autoencoder trained to produce disentangled vectors of content, emotion, and style, 
(2) a 3D body motion prior in the form of a temporal variational autoencoder (VAE) to generate smooth and realistic gestures, and
(3) a latent diffusion model, which generates 3D body motion given the input content, emotion, and style latent vectors.

Training such a model requires a speech-to-3D body dataset of sufficient scale, which is rich and diverse in speakers and emotions.
BEAT \cite{liu2022beat} is a good candidate because it provides a large set of 3D gestures associated with single-person monologues.
Unfortunately, the bodies are represented as skeletons\rebut{, and it lacks face mocap markers and FLAME expressions.}
Instead, to produce realistic body animations, we require articulated 3D body surfaces.
To overcome this, we convert BEAT sequences to SMPL-X \cite{SMPL-X:2019} format using MoSh++ \cite{AMASS:2019} and use the SMPL-X parameters for training.
\rebut{See \cite{liu2024emage} for comparison.}

Our contributions are:
(1) We present a framework to synthesize emotional 3D body gestures directly from speech.
(2) We factor an input audio into disentangled content, emotion and style vectors, which enables us to separately control emotion in generated gestures. 
(3) We adapt temporal latent diffusion for multiple target conditions.

\section{Related Work}
\label{sec:relatedworks}

\subsection{3D Conditional Human Motion Generation}

Early works focus mostly on predicting~\cite{Ormoneit2005RepresentingCH, Yuan2020DLow, Corona2020ContextAwareHM, Barsoum:2017, Martinez:2017, Salzmann2022MotronMP, Liu2022TowardsRA,Harvey2020RobustMI, Zhou:2020, Kim2022ConditionalMI} or generating human motion~\cite{Habibie2017rnn, Li2022GANimatorNM}, but do not consider  multi-modal control. Recently, conditional motion generation through other modalities, such as text~\cite{Ahuja2019Language2PoseNL,Ghosh_2021_ICCV,petrovich22temos,TEACH:3DV:2022,Guo2022GeneratingDA, dabral2023mofusion, SINC:2023}, music~\cite{Li2021LearnTD, Moltisanti2022BRACETB,tseng2023edge}, speech~\cite{Habibie2022AMM}, or action labels~\cite{chuan2020action2motion,ACTOR:ICCV:2021}, has gained more attention. 
Below, we focus on speech-driven motion generation methods, since they are the most relevant to our work.

\subsection{Gesture Generation from Speech}

\paragraph{Rule-based gesture synthesis.} Embodied conversational agents (ECA) are designed to interact and communicate with humans. Using the Behavior Markup Language (BML) \cite{10.1007/11821830_17} one can build rule-based systems for humanoids based on predefined behaviors~\cite{Poggi2005}. 
This is used for completion of a storytelling task in an expressive manner~\cite{6100857}. 
The BEAT rule-based toolkit~\cite{Cassell2004} enables adding non-verbal behavior on top of a pre-animated figure. Thiebaux et al.~\cite{inproceedingssmartbody} develop an ECA by using procedural animation techniques and keyframe interpolation. Marsella et al.~\cite{10.1145/2485895.2485900} design a generalized rule-based agent to generate expressions, eye gaze, and gestures from speech. Each of these approaches are based on non-trainable, rule-based techniques that may require substantial manual modelling effort to adapt to new tasks.  

\paragraph{Data-driven gesture synthesis.} More recently, data-driven methods have superseded rule-based systems. 
Yoon et al.~\cite{Yoon2020Speech} use a fusion of text, audio and upper body gestures to learn an upper body gesture avatar, but  can only control the style of individual speakers by sampling from their latent space. SpeechGestureMatching~\cite{Habibie2022AMM} generates 3D facial meshes and 3D keypoints of the body and hands from speech, but the outputs are separated and the method does not provide  control over the generations. QPGesture~\cite{yang2023QPGesture} uses phase to better align the generated 3D skeleton-based gesturing avatars with the audio input. Ginosar et al.~\cite{ginosar2019gestures} 
and Diverse-3D-Hand-Gesture-Prediction~\cite{qi2023diverse} generate hand and arm motions only. Audio2Gestures~\cite{li2021audio2gestures} encode motion and audio to a low-dimensional latent space and generate gestures.  SEEG~\cite{liang2022seeg} aims to generate gestures that align well with the semantics of the speech. Diff-TTSG~\cite{mehta2023diffttsg} regresses speech and gestures at the same time, joining the two modalities in a single system. DiffGAN~\cite{Ahuja_2022_CVPR} retargets gestures across speakers in a low-resource setting. The GENEA challenge~\cite{yoon2022genea} tackles gesticulation from speech alone using the Talking-with-Hands dataset~\cite{lee2019talking}. Gesture2Vec~\cite{9981117} uses a machine translation model to translate text into gesture chunks and output full sequences using such quantized representations. TalkSHOW~\cite{yi2023generating} uses a VQ-VAE to generate 3D human bodies gesturing with facial expressions from speech segments, but in an uncontrolled manner. Similarly, Co-speech gesture \cite{lu2023cospeech} uses an RQ-VAE to generate different gestures from speech.
Alternative gesture generation from speech methods have been proposed such as reinforcement learning~\cite{racer}, self-supervised pre-training~\cite{10095344}, and diffusion~\cite{Zhu_2023_CVPR, mehta2023diffttsg}. 
BodyFormer~\cite{10.1145/3592456} introduces a dataset of pseudo-groundtruth and a transformer-based method for generating gestures from speech. However, none of these methods provide explicit emotional control over the generated motion.

For controllable generation, GestureDiffuCLIP~\cite{Ao2023GestureDiffuCLIP}\rebut{ incorporates multiple conditions including} CLIP~\cite{radford2021learning} text features\rebut{, video, or motion prompts via AdaIn~\cite{huang2017arbitrary} layers} to generate gestures from speech, \rebut{however, it does not allow explicit control over the emotion conveyed by the driving audio.} ListenDenoiseAction~\cite{alexanderson2023listen} combines conformers and the DiffWave~\cite{kong2021diffwave} architecture to generate gestures that can be controlled by a style vector, RhythmicGesticulator~\cite{Ao_2022} disentangles the latent space into a vector related to the semantics of the gesture and one related to the subtle variations, while DisCo~\cite{liu2022disco} models content and rhythm. StyleGestures~\cite{alexanderson2020style} adapts MoGlow~\cite{henter2020moglow}, demonstrating limited control over some motion attributes like the speed and expressiveness of gestures. DiffuseStyleGesture~\cite{yang2023DiffuseStyleGesture} uses diffusion to generate diverse gestures from speech.

\subsection{Emotion Control}
Emotion classification and control has been little studied  in 3D human motion generation
with only a a few methods using skeletal motion in multi-class classification.
Ghaleb et al.~\cite{ghaleb2023cospeech} employ a spatio-temporal graph convolution network to classify gestures into four classes: preparation, stroke, retraction, and neutral.
Li et al.~\cite{Li2016Affectcla}, on the other hand, use hidden Markov models for emotion classification of human movement mocap data.
Karras et al.~\cite{karrasemotion} learn face animations of a single actor, and test their method on different tasks by modifying the latent vectors. However, there is no disentanglement mechanism, and they do not model the synchronization of the emotion with the with the facial motions. Recently, EmoTalk~\cite{peng2023emotalk}, animates emotional 3D faces from speech input with control over the emotion intensity and EMOTE~\cite{danecek2023emotional} disentangles emotion and speech to allow emotion editing at test time. However, models solely intended for facial tasks like lip syncing and capturing expressions might not smoothly adapt to the complexity of whole-body movements and distinct articulation.
Regarding emotion-conditioned motion generation, Aberman et al.~\cite{aberman2020unpaired} show style-transfer from video data to motion and provide some style-based control, but do not address speech-driven emotional gestures.
Similarly, the ZeroEGGs~\cite{ghorbani2022zeroeggs} dataset contains some emotional gesture controls but also includes more generic styles of motion. The method requires the input of arbitrary frames of desired motion to encode a style, thereby relying on motions and speech as conditions during inference.
\rebut{Text-driven emotional gesticulation, as explored by Bhattacharya et al.~\cite{DBLP:conf/vr/BhattacharyaRBG21,DBLP:conf/mm/BhattacharyaCRM21}, emphasizes the generation of gestures based on textual cues, incorporating additional conditions such as speech, speaker ID, seed poses, as well as valence, arousal, and dominance triplets. However, these approaches do not provide the means to distill explicit emotion features, limiting free control over the generated gestures.} 
Closer to our work, EMoG~\cite{yin2023emog} incorporates emotion cues from the BEAT dataset~\cite{liu2022beat} to generate improved gesture quality without explicit emotion control.
EmotionGesture~\cite{qi2023emotiongesture} uses a TED Emotion Dataset and BEAT to incorporate emotion features in gesture generation and generate emotional gestures. Although they can generate emotional gestures, their method is not end-to-end and has no explicit motion control. 
Specifically, it uses an emotion-conditioned VAE after training to acquire diverse emotion features that are used to generate gestures without guarantees and control over emotion types. Wu et al.~\cite{Wu2023} introduce the first multi-cultural gesture dataset containing 200 individuals of 10 different cultures. 
In contrast to prior work, we explicitly control the emotions conveyed by the generated gestures solely through emotional speech \rebut{without relying on additional conditions.}

\begin{figure*}[t]
    \centering
    \captionsetup{type=figure}
    \includegraphics[width=1\textwidth]
    {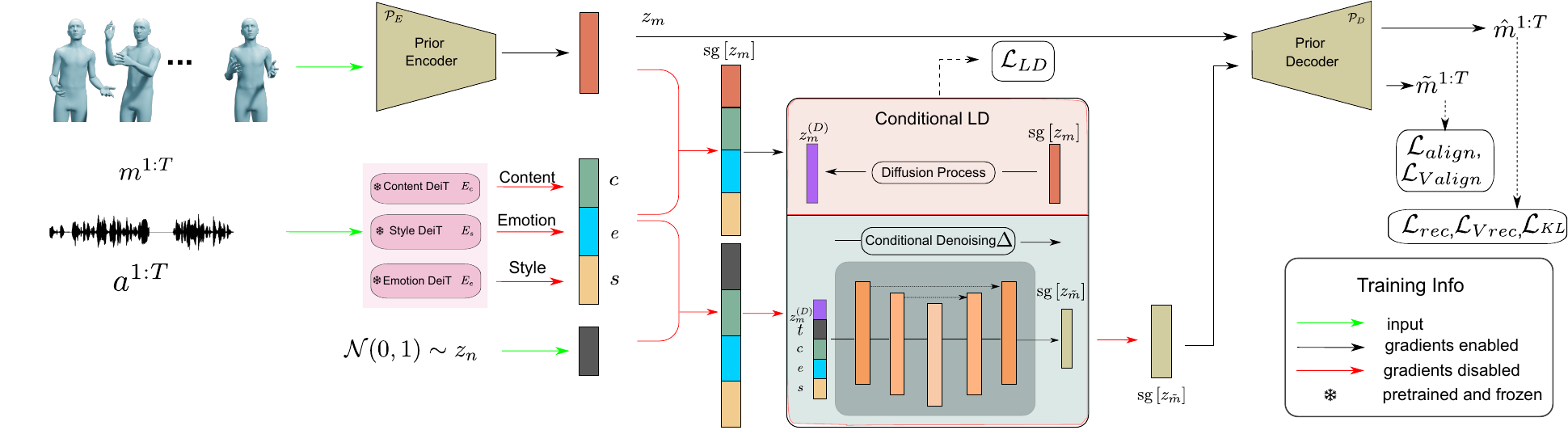}
    \captionof{figure}{
    \textbf{Training.} 
    We train the motion prior ($\priorenc,\priordec$) and the latent denoiser $\denoisernet$ jointly, while keeping the audio encoding networks frozen. In the forward pass, we take an input audio $\mfcc\tmp$ and pose sequence $\motion\tmp$. 
    Firstly, we do a forward pass of  $\motion\tmp$ through $\priorenc$ and $\priordec$ and compute $\mathcal{L}_{rec}$, $\mathcal{L}_{Vrec}$, and $\mathcal{L}_{KL}$.
    Then, we apply the diffusion process to a gradient-detached $\sg{\latentmotion}$ obtaining the noisy $\latentmotion^{(D)}$, which is then denoised with $\denoisernet$ and $\mathcal{L}_{LD}$ is computed.
    Finally, we use $\denoisernet$ to fully denoise $\normalvec$ into gradient-detached $\sg{\denoisedlat}$, further decode  $\diffdecodemotion$ using $\priordec$, and compute  $\mathcal{L}_{align}$ and $\mathcal{L}_{Valign}$.
    }
    \label{fig:arch}
\end{figure*}

\section{Method}
\label{sec:method}

The \modelName pipeline consists of two separately trained networks. The audio disentanglement module, which encodes input speech into latent vectors for content, emotion, and style is described in \cref{subsec:disentanglemodel}.
The main architecture is described in  \cref{subsec:gesturemodel}.
It consists of a 3D human motion prior coupled with a latent diffusion model. 
It takes random noise (or partially denoised latent vectors) on the input and outputs a human motion sequence. 
We introduce broader applications in gesture editing in \cref{subsec:apps}.

\subsection{Preliminary: Expressive 3D Body Model}

\smplx~\cite{SMPL-X:2019} is a \threeD model of the body surface. 
\smplx is defined as function $\mesh(\shape, \pose, \expression)$ that produces a \threeD body mesh. 
It is parameterized by identity shape $\shape \in \mathbb{R}^{300}$, pose $\pose\in \mathbb{R}^{J\times 3}$ including finger articulation for rotations around $J$ joints, and facial expression $\expression \in \mathbb{R}^{100}$. We adopt the continous 6D rotation representation for training following Zhou et al.~\cite{zhou2020continuity}, making $\pose\in \mathbb{R}^{J\times 6}$. Given pose parameters and any shape parameter, we can obtain body mesh vertices $\vertices$ using the differentiable \rebut{SMPL-X layer}~\cite{SMPL-X:2019}.
As the focus of our paper is on synthesizing body gestures and not locomotion, we disregard 8 joints that correspond those of the lower body joint poses, leaving $J=\numjoints$.
Further, we omit the facial expression parameters, i.e., set $\expression = \mathbf{0}$.

\subsection{Speech Disentanglement Model}
\label{subsec:disentanglemodel}
\paragraph{Architecture.}
The goal of the this model is to factor an input speech into three disentangled latent representations, one for content (i.e., the words spoken), one for emotion, and one for personal style. 
To do so, we devise a specialized encoder--decoder architecture 
with three separate encoders, one for each latent space.
We denote the encoders as: $\enccontent(\mfcc) = \tmplatentcont, \mkern27mu \encemotion(\mfcc) = \tmplatentemo, \mkern27mu \encstyle(\mfcc) = \tmplatentstyle$,
where $\mfcc$ is the input filterbank, $\tmplatentcont$, $\tmplatentemo$ and $\tmplatentstyle$ denote the latent vectors for content, emotion and style and $\enccontent$, $\encemotion$ and $\encstyle$ are their encoders. 
The architecture of the three encoders follows the design by Gong et al.~\cite{gong2021ast, gong2022ssast} (i.e., leveraging the DeiT visual transformer \cite{TouvronCDMSJ21deit} adapted for processing filterbank images extracted from the input audio). 
The decoder takes the three latent vectors and produces a reconstructed filterbank. Formally $\dec(\tmplatentcont, \tmplatentemo, \tmplatentstyle) = \widehat{\mfcc}$,
where $\widehat{\mfcc}$ denotes the reconstructed filterbank. 
The decoder architecture consists of a fusion module and transformer-encoder layers.

\paragraph{Training.}
The audio module is trained with a multiple loss terms that ensure that the three latent spaces are properly disentangled.
In addition to the standard autoencoder reconstruction loss, we also employ three cross-reconstruction losses, in which we enforce the correct reconstruction of the audio signal where we modify one of the content, style or emotion latents. 
Additionally, we employ three loss terms on the latent vector predictions -- namely emotion and style classification losses over $\tmplatentemo$ and $\tmplatentstyle$, and a content similarity loss between 
pairs of two content latent vectors extracted from audios that have the same spoken content.
For a detailed description of the encoder--decoder architecture, a formal definition of the loss functions and a detailed description of the training process please refer to the \supmat

\subsection{Gesture Generation Model}
\label{subsec:gesturemodel}

\paragraph{Motion prior.}
Similar to~\cite{chen2023executing,petrovich21actor}, our motion prior network is a VAE transformer architecture with encoder $\priorenc$ and decoder $\priordec$.
Specifically, both $\priorenc$ and $\priordec$ follow a U-Net-like \cite{ronneberger2015unet} structure with skip connections between transformer blocks (see \supmat~for details).
The positional embeddings are learnable and injected into each multi-head attention layer, following the design of Carion et al.~\cite{carion2020endtoend}. 
Formally, the encoder takes a sequence of $T$ frames of the SMPL-X pose vectors $\motion\tmp \in \mathbb{R}^{6\joints \times T}$ and the first two tokens of its output, $\mu \in \mathbb{R}^{\motiondim}$ and $\Sigma \in \mathbb{R}^{\motiondim \times \motiondim}$ are used to extract the motion latent 
$\latentmotion \in \mathbb{R}^{\motiondim}$ via the reparametrization trick. The decoder takes zero positional encodings as query input and the motion latent is fed as memory to every cross-attention transformer layer, producing the reconstructed motion $\priordecodemotion$.

\paragraph{Diffusion process.} 
The forward diffusion process is similar to~\cite{ho2020denoising, nichol2021improved}. We employ fixed variance and linearly scaled noise scheduler. 
We add noise to the motion latent $\latentmotion$ for $D$ diffusion timesteps to obtain $z^{(D)}$ following:
\begin{equation*}
    q(\latentmotion^{(\difft)} \mid  \latentmotion^{(0)}) = \mathcal{N}(\latentmotion^{(\difft)} ; \sqrt{\bar{\alpha }_{\difft} }\latentmotion^{(0) },(1-\bar{\alpha }_{\difft})\textbf{I}),
\end{equation*}
with $\alpha _{\difft} = 1-\beta _{\difft}$, $\bar{\alpha }_{\difft} = \prod_{s=1}^{\difft}{\alpha }_{s}$, and $\beta _{\difft}$ denotes diffusion process variance. 

\paragraph{Conditional denoising process.}
The denoising process consists of iteratively denoising a conditioned noisy motion latent vector to obtain the denoised motion latent  $\diffdecodemotionlatent$. 
Our denoiser $\denoisernet$ is a latent variable model \cite{rombach2021highresolution}
and its architecture is similar to the U-Net-like structure of the motion prior encoder $\priorenc$. 
The input of the model is a concatenation of: $\latentmotion^{(\difft)}, \pe{\difft},\tmplatentcont , e,\tmplatentstyle \in \mathbb{R}^{\latentdim}$, where $\pe{\difft}$ is a sinusoidal positional encoding of diffusion timestep $\difft$ as defined in \cite{ho2020denoising}. 
$\denoisernet$ iteratively denoises through each reversed diffusion step:
\begin{equation*}
\latentmotion^{(\difft-1)} = \denoiser([ \latentmotion^{(\difft)},\pe\difft,c,e,s]) 
.
\end{equation*}

\paragraph{Training.} 
We optimize the motion prior and the latent denoiser jointly  to ensure audio--motion latent code alignment during conditional fusion in the denoising process using a 3-step forward pass through the gesture generation model. 
First, following standard VAE practice, we reconstruct $\hat{\motion}\tmp$ by the motion prior forward pass. 
As shown in Fig.~\ref{fig:arch}, we then disable gradient calculation in $\priorenc$ to infer the intermediate motion latent $\sg{z_{m}}$, which serves as input to the denoiser. At this stage, we obtain the denoiser noise prediction, $\denoiserpred$ and use to compute the diffusion model gradients. 
Finally, in the third step we compute $\tilde{\motion}\tmp = \priordec(\sg{z_{\tilde{m}}})$, where $z_{\tilde{m}}$ is obtained by iteratively using the $\denoiser$ to obtain a fully denoised latent from $z_n^{(t_D)} \sim \gaussian$. 
We indicate computations done without gradients with a stop-gradient operation $\sg{.}$. 

\paragraph{Losses.}
\label{subsec:gesturemodellosses}
To train the motion prior, we include the standard VAE losses, namely the reconstruction loss on pose parameters $ \mathcal{L}_{rec}$ and on vertex coordinates $ \mathcal{L}_{Vrec}$ using the smooth L1 metric introduced in \cite{girshick2015fast}, which we denote as $\smoothL$:
\begin{equation*}
     \mathcal{L}_{rec} = \smoothL(\motion\tmp,\hat{\motion}\tmp),  \mkern18mu
  \mathcal{L}_{Vrec} = \smoothL(\vertices\tmp,\hat{\vertices}\tmp),
\label{equ:rec}
\end{equation*}
where the root-centered vertices $\vertices$ are obtained by feeding in pose parameters $\motion$ to a differentiable SMPL-X layer (without learnable parameters) and a mean shape $\shape=\vec{0}$. The KL divergence loss of the motion prior is:
\begin{equation*}
    \mathcal{L}_{KL} = \frac{1}{2} \left[ \sum_{i=1}^{z}(\mu_{i}^{2}+\sigma_{i}^{2})  - \sum_{i=1}^{z} \left(log(\sigma_{i}^{2}) + 1 \right) \right].
  \label{eq:kld}
\end{equation*}

To ensure the alignment of the diffusion-generated motions and the input audio, we apply the alignment reconstruction loss on the inferred motion pose parameters and the vertex coordinates:
\begin{equation*}
   \mathcal{L}_{align} = \smoothL(\motion\tmp,\tilde{\motion}\tmp), \mkern18mu
 \mathcal{L}_{Valign} = \smoothL(\vertices\tmp,\tilde{\vertices}\tmp).
\label{equ:align}
\end{equation*}

Finally, we utilize the objective similar to \cite{rombach2021highresolution, chen2023executing,ho2020denoising} to supervise the denoiser:
\begin{equation*}
    \mathcal{L}_{LD} =  \left \| \denoiserpred^{(\difft)} - \denoiser (z^{(\difft)}_m,\pe{\difft},\tmplatentcont,\tmplatentemo,\tmplatentstyle)\right \|^{2}_{2},
  \label{eq:denoising}
\end{equation*}
where $\denoiserpred^{(\difft)}$ is the noise vector sampled from $ \mathcal{N}(\mathbf{0},\mathbf{I}) $ in the corresponding diffusion step $\difft$.
The combined gesture model loss is:
\begin{equation*}
\mathcal{L}_{ges} =\mathcal{L}_{rec} + \mathcal{L}_{Vrec}+ \mathcal{L}_{KL}+ \mathcal{L}_{align} +\mathcal{L}_{Valign}+ \mathcal{L}_{LD}
  \label{eq:gesturecombinedloss}
\end{equation*}

\paragraph{Inference.}
We employ DDIM~\cite{song2020denoising} to infer high quality conditional motion samples 
with a small number of denoising timesteps. 
During inference we draw a sample vector from $\gaussian$ to iteratively denoise in reversed timesteps.
The denoised sample is then passed through the decoder $\priordec(\diffdecodemotionlatent)$ to obtain motion $\tilde{\motion}\tmp $ .
\subsection{Gesture Editing}
\label{subsec:apps}
Due to the disentangling of the inputs, \modelName achieves semantic gesticulation control using two driving input audios.
Specifically, given two input audio signals $\mfcc_1$ and $\mfcc_2$, we extract their latent representations of content $\tmplatentcont_1, \tmplatentcont_2$, emotion $\tmplatentemo_1, \tmplatentemo_2$, and style $\tmplatentstyle_1, \tmplatentstyle_2$. 
Then, we simply initialize the denoising procedure of $\denoiser$ with the triplet
$(\tmplatentcont_1,\tmplatentemo_2,\tmplatentstyle_1) $, 
generating the gesture with the content and style of $\mfcc_1$ but the emotion of input audio $\mfcc_2$.
Similarly, instead of emotion we can also change the gesticulation style to that of the speaker of $\mfcc_2$ by initializing with $(\tmplatentcont_1,\tmplatentemo_1,\tmplatentstyle_2) $.

\section{Implementation Details}
\label{sec:implementation}
\paragraph{MoCap data preparation.}
The BEAT~\cite{liu2022beat} mocap sequences, captured in a Vicon system at 120 Hz, are downsampled to 30 Hz and processed using MoSh++~\cite{Loper:SIGASIA:2014, AMASS:2019} to obtain SMPL-X parameters. 
Given a sequence of 3D mocap marker positions, we jointly optimize SMPL-X shape and pose parameters, 3D body translation, and embedding of the mocap markers in the SMPL-X surface.
Once processed, the sequences are then divided according to the emotion annotations in the BEAT dataset. 
We use sequences of English speaking subjects in monologue speaking style for training and evaluating \modelName.
For each sequence we draw $\motion\tmpfull$ at 30 FPS and concatenate with audio content $c$, emotion $e$, and style $s$ latent vectors. Then, we segment it to 10-sec windows $T$, beginning from the timestamp 0 and discarding additional unaligned information at the end. 
This preprocessing choice allows us to train transformer networks without masking. 
We provide additional data processing information in the \supmat

\paragraph{Audio preprocessing.}
We use audio sequences belonging to eight categorical emotion labels (neutral, happy, angry, sad, contempt, surprise, fear, and disgust). 
Each audio chunk of 10s is converted into a filter bank with 128 mel-frequency bins with a 25ms Hamming frame window and 10ms frame shift. 
We mask each sample with a maximum length of 24 in the frequency domain and a maximum length of 96 in the time domain, employing Park et al.~\cite{Park_2019}. 
Following \cite{gong2021ast, gong2022ssast}, we standardize the filter bank and augment it via noise injection and circular shifting. 
Before feeding in our speech disentanglement model, each filter bank is split into a sequence of fixed 1209 patches of  16 x 16  each having 6 units overlap in frequency and time domain.
 
\paragraph{Motion prior.} 
The motion prior is a VAE encoder--decoder with 9 layers and 4 heads, following Chen et al.~\cite{chen2023executing}. The encoder--decoder is a U-Net-like transformer with residual connections. Learnable positional embeddings are injected in each multi-head attention layer. We have a linear projection at the start and the end of our motion prior network. The KL divergence term is weighted with a factor of $1e-4$. 

\paragraph{Denoiser.} The denoiser follows the same network architecture as our prior encoder. 
The hidden dimension of all transformer layers is 1024.
We use 1000 diffusion steps $D$ during training and 50 during inference. 
Noise betas are in range $[0.00085,0.012]$. 
We jointly optimize the prior and denoiser networks for 5000 epochs with batch size of 64, learning rate $0.0001$, and the AdamW optimizer \cite{loshchilovH19adamw}.

\section{Experiments}
\label{sec:experiments}

\paragraph{Speech disentanglement model.} 
We evaluate the performance of the speech disentanglement model quantitatively using classification accuracy and F1 scores on emotion and style. 
The accuracy is computed as average scores for all 8 emotion as well style categories that are part of the test dataset. 
The emotion and style accuracy is 91.53\% and 96.06\%, respectively. 
The emotion F1 score and style F1 scores are 0.914 and 0.960, respectively. 
See the \supmat~for ablations and a detailed metric analysis.

\paragraph{Gesture generation model.} 
We evaluate the performance of our gesture generation model quantitatively, qualitatively, and perceptually against following methods: TalkSHOW \cite{yi2023generating} and the re-implementation of Habibie et al.~\cite{3dconvgesture_2021} provided by the TalkSHOW authors in the official TalkSHOW release~\cite{showgithub}, DiffuseStyleGesture (DSG)~\cite{yang2023DiffuseStyleGesture}, MoGlow~\cite{henter2020moglow}, and CaMN~\cite{liu2022beat}. Additionally, we adapt TalkSHOW  to include categorical emotion labels as input along with the existing architecture that only allows one-hot encodings of personal style. We then retrain it on our training data. We refer to it as TalkSHOW-BEAT. There are some concurrent works~\cite{Ao2023GestureDiffuCLIP,alexanderson2023listen,liu2024emage}, which introduce methods for gesture generation from speech, however, direct comparison is hindered by the unavailability of released code our task. Refer to the \supmat~for the ablation experiments, the emotion and style editing experiments, and their quantitative evaluation. 
\subsection{Quantitative Evaluation}
\label{sec:quantexperiments}

\begin{table}[b]
\centering
\scalebox{0.85}{
\begin{threeparttable}
\begin{tabular}{llllll}\centering
Method & SRGR$\uparrow$ & BA$\uparrow$  & FGD$\downarrow$  & Div$\rightarrow$  & GA\tnote{a}\;$\uparrow$ \\ 
\midrule
    GT  & ---& 0.83 & --- & 27.83& 64.04  \\
\midrule
    Ours & \cellcolor{green!25}0.36& \cellcolor{green!25}0.81 & \cellcolor{green!25}388.63 & \cellcolor{green!25}25.06& \cellcolor{green!25}46.76  \\
    Ours-EmoEdit\tnote{b} & ---& \cellcolor{blue!25}0.79 & 792.58 & \cellcolor{blue!25}24.68& \cellcolor{blue!25}34.18 \\
    TalkSHOW-BEAT & \cellcolor{blue!25}0.31& 0.64 & 808.99 & 24.16& 22.71 \\
    TalkSHOW \cite{yi2023generating} & 0.30& 0.60 & \cellcolor{blue!25}762.15 & 23.19& 29.41   \\
    DSG~\cite{yang2023DiffuseStyleGesture} & 0.23& 0.40 & 763.10 & 19.77& 22.70 \\
    Habibie et al.~\cite{3dconvgesture_2021} & 0.23& 0.39 & 809.17 & 21.34& 16.67 \\
    MoGlow~\cite{henter2020moglow} & 0.21& 0.35 & 1097.03 & 19.50& 16.62 \\
    CaMN~\cite{liu2022beat} & 0.21& 0.39 & 1063.87 & 18.90& 14.17 \\
    
\bottomrule
\end{tabular}
\begin{tablenotes}

            \item[a] GA is average of all 8 emotions. 
            \item[b] GA for these are average accuracy for all generations with 7 edited audio sequences.
        \end{tablenotes}
\end{threeparttable}
}
\caption{\textbf{Gesture quantitative results.} We compare our methods against several SOTA methods using metrics explained in~\cref{sec:quantexperiments}. We observe that \modelName outperforms in all scores compared to baseline methods. Additionally, \modelName-EmoEdit outperforms in Beat Align, Diversity, and Gesture Emotion Accuracy scores compared to the baseline methods.
}
 \label{tab:quant}
\end{table}

To quantitatively evaluate our method’s gesture generations and edited gesture generations, we train a transformer-based encoder architecture (denoted as $\mfeatxtractor$) similar to Petrovich et al.~\cite{petrovich21actor} in an autoencoder setting, where we append a $\textup{CLS}$ token at the beginning of the motion sequence. $\mfeatxtractor$ is trained with a cross-entropy emotion classification objective applied to the output $\textup{CLS}$ token. We train $\mfeatxtractor$ on the BEAT training dataset and use its features to compute the following metrics:
(1) Fréchet gesture distance (FGD): We follow \cite{yoonICRA19, Yoon2020Speech, Seitzer2020FID} to compute the feature distance between generated and ground truth motion features. 
(2) Gesture diversity (Div): Similarly to Chen et al.~\cite{chen2023executing}, we compute variance across generated features.
(3) Gesture emotion accuracy (GA): We report top-1 emotion classification accuracy predicted by a classifier trained on the motion $\mfeatxtractor$-predicted latents. 
(4) Beat align (BA): We follow~\cite{li2021ai, liu2022beat}, to evaluate the motion-speech correlation in terms of the similarity between the kinematic motion beats and speech audio beats. The kinematic motion beats are directly computed from the generated motion sequences. 
(5) Semantic-Relevant Gesture Recall (SRGR): We follow Liu et al.~\cite{liu2022beat}, to evaluate the semantic relevancy of gestures with GT motion. We use the ground truth semantic scores to compute this metric. The scores are obtained from the BEAT authors,  representing a continuous score on a scale 0-1 per gesture style for 4 gesture semantic categories: beat, deictic, iconic, and metaphoric.
While comparing with methods that output coarse skeletal data (DSG~\cite{yang2023DiffuseStyleGesture}, MoGlow~\cite{henter2020moglow}, and CaMN~\cite{liu2022beat}), we convert the skeleton motion data into the SMPL-X axis angle representation.
For details on the architectures and training of $\mfeatxtractor$, and the losses, please refer to the \supmat

We prepare the evaluation data by randomly selecting 72 unique motion sequences each of length 10s and comprising 8 emotions across test subjects and compute the aforementioned metrics. We use 9 sequences for each emotion per subject. The results are reported in Tab.~\ref{tab:quant}. All best scores are highlighted in \colorbox{green!25}{green} and second best in \colorbox{blue!25}{blue}.
\modelName outperforms the baseline methods in all given metrics. 
\rebut{
To validate the performance of gesture emotion editing, we also report the same metrics for the emotion editing task (Ours-EmoEdit). 
During inference, the input style and content latents are extracted from neutral-emotion audio, while the emotion latent comes from a different audio 
of different
emotion. These emotional edits offer numerous possibilities, allowing for transitions from any to any emotion. Tab.~\ref{tab:quant} shows the average for 
editing from neutral to other emotions.
}
Since we require the GT gesture semantics score to compute SRGR metric, it is not possible to compute the SRGR for the synthetic edited-emotion gestures as they are not part of the original BEAT dataset. Ours-EmoEdit outperforms the baseline methods in BA, Div, and GA metrics. This demonstrates the capability of our model to maintain highly discriminative cues when switching between different emotions. TalkSHOW-BEAT has the second best score for SRGR whereas TalkSHOW demonstrates second best FGD score. 
Although, our model and ours-EmoEdit show improvements over the baseline methods, GT motions have higher diversity, Beat alignment score, and are easier to classify than generations of \modelName, highlighting the challenging nature of the problem.

\begin{figure*}[t!]
    \centerline{
    \includegraphics[width=1\textwidth]
    {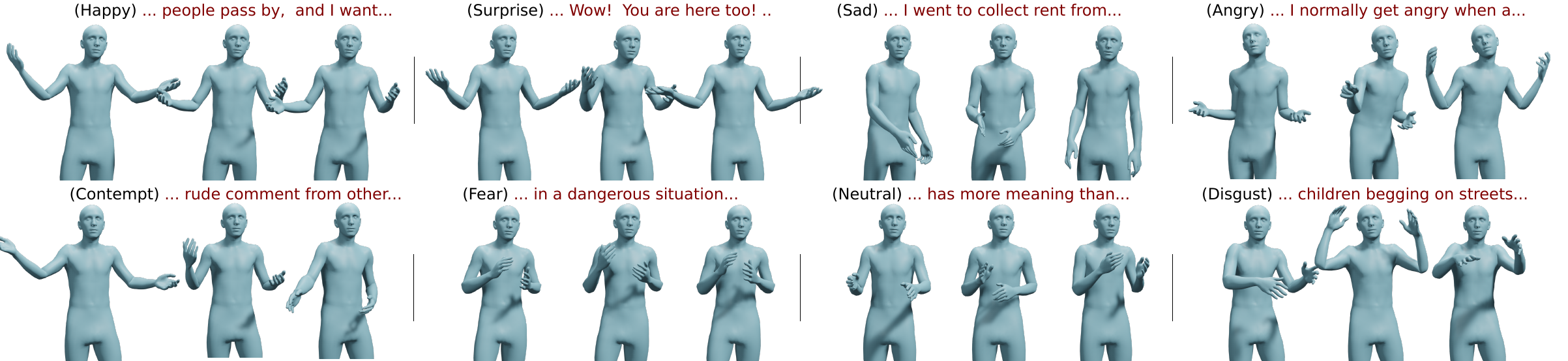}
    }
    \vspace{-0.08in}
    \caption{
    \textbf{Qualitative comparison across all emotions.}
    We evaluate generation on different test audios.  \modelName exhibits well-synchronized beat gestures and consistently produces gestures that accurately convey the emotional content expressed in the input speech.
    }
    \label{fig:gesturesem}
\end{figure*}

\begin{figure}[t]
  \centering
  \includegraphics[width=1\linewidth]{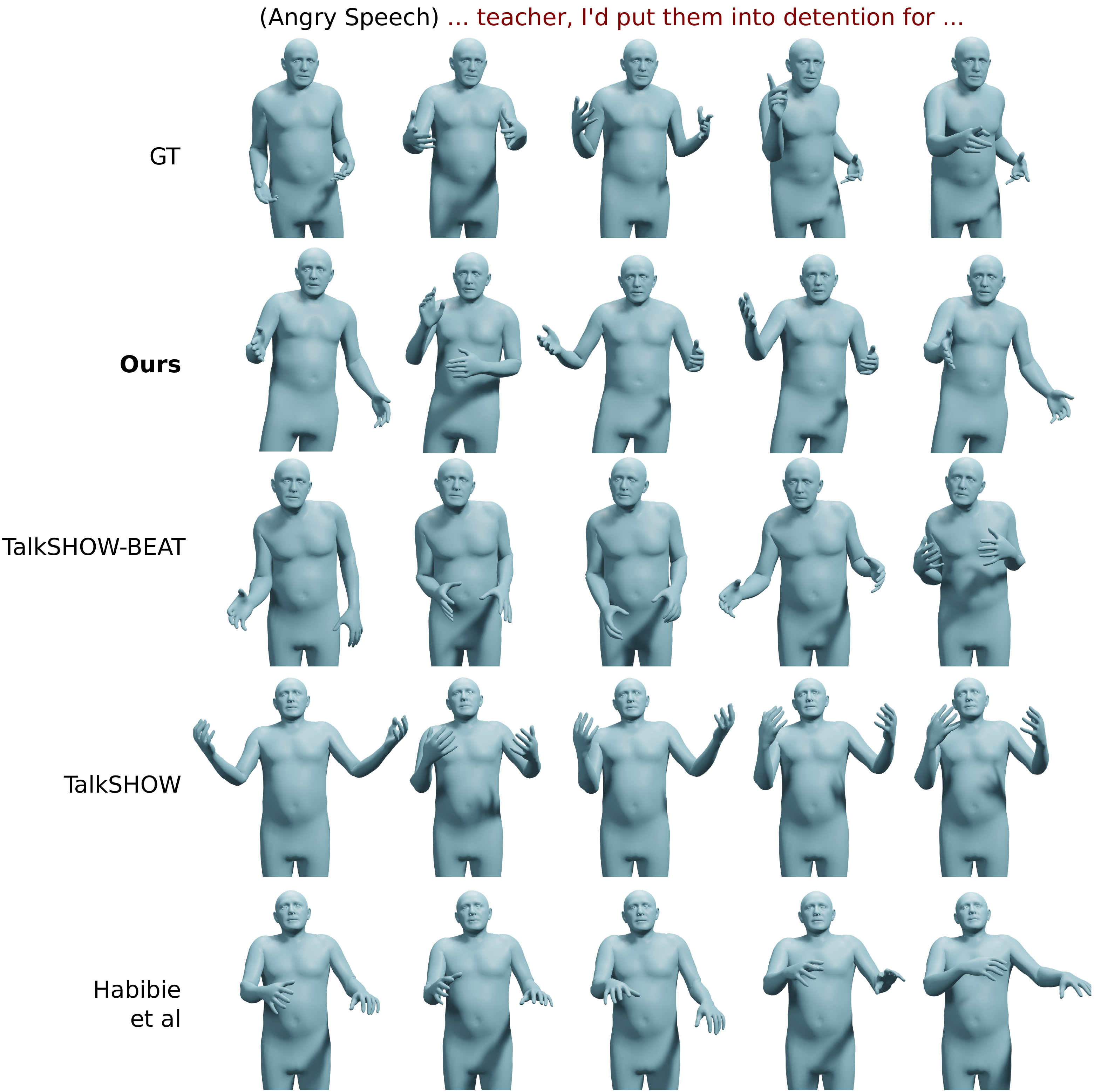}

   \caption{
   \textbf{Qualitative comparison with baseline methods}. The speech segment describes intense angry speech.
   }
   \label{fig:baselines}
\end{figure}

\subsection{Qualitative Evaluation}
\label{sec:qualexperiments}

\paragraph{Comparison with baseline methods.}
In \cref{fig:baselines}, we demonstrate comparison with baseline methods that output a 3D body mesh: Habibie et al.~\cite{3dconvgesture_2021}, TalkSHOW \cite{yi2023generating}, TalkSHOW-BEAT, and the BEAT ground truth (GT) \cite{liu2022beat}. 
We observe that \modelName generates gestures that are semantically closer to the speech content and produces expressive emotional gestures closer to the perceived emotion. For example, the GT motion exhibits anger when saying ``\red{\textit{put them into detention}}''. \modelName demonstrates tense posture and aggressive movements comparable with the ground truth data and accurate synchronization with the spoken words. TalkSHOW \cite{yi2023generating} and Habibie et al.~\cite{3dconvgesture_2021}  exhibit limited movement and display inferior and static gestures on test audios as seen in the last two rows of~\cref{fig:baselines}. TalkSHOW-BEAT slightly outperforms other baseline methods by demonstrating enhanced synchronized gestures, but it still does not perform as well as \modelName.

\begin{figure}[t]
  \centering
  \includegraphics[width=0.75\linewidth]{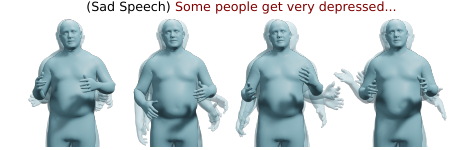}

   \caption{  \textbf{Qualitative evaluation of diverse generations}. Multiple generations overlayed.}
   \label{fig:diversity}
\end{figure}

\paragraph{Diverse emotional gestures.}
In \cref{fig:diversity}, our probabilistic model can generate diverse gestures for same input audio.

\paragraph{Emotional gesture generation.} In \cref{fig:gesturesem} \modelName demonstrates strong correlation with the spoken utterances as well as different emotions. We observe that our model is able to correlate semantic words to associated gestures. For example, gestures demonstrate forceful actions and tense stance with angry audio ``\red{\textit{normally get angry}}'' whereas it generates lowered and calm hand positions for sad audio ``\red{\textit{I went to collect}}''. Similarly, our generations show hands that are closer to body for fearful audio ``\red{\textit{in a dangerous situation}}'' while widely open expressing astonishment for happy and surprised audio ``\red{\textit{people pass by}}'' and ``\red{\textit{Wow! You are here}}''.

\begin{figure*}[t]
  \centering
  \hspace{-1cm}
    \includegraphics[trim={0 2.5cm 4.5cm 2.5cm},clip, scale=0.35]{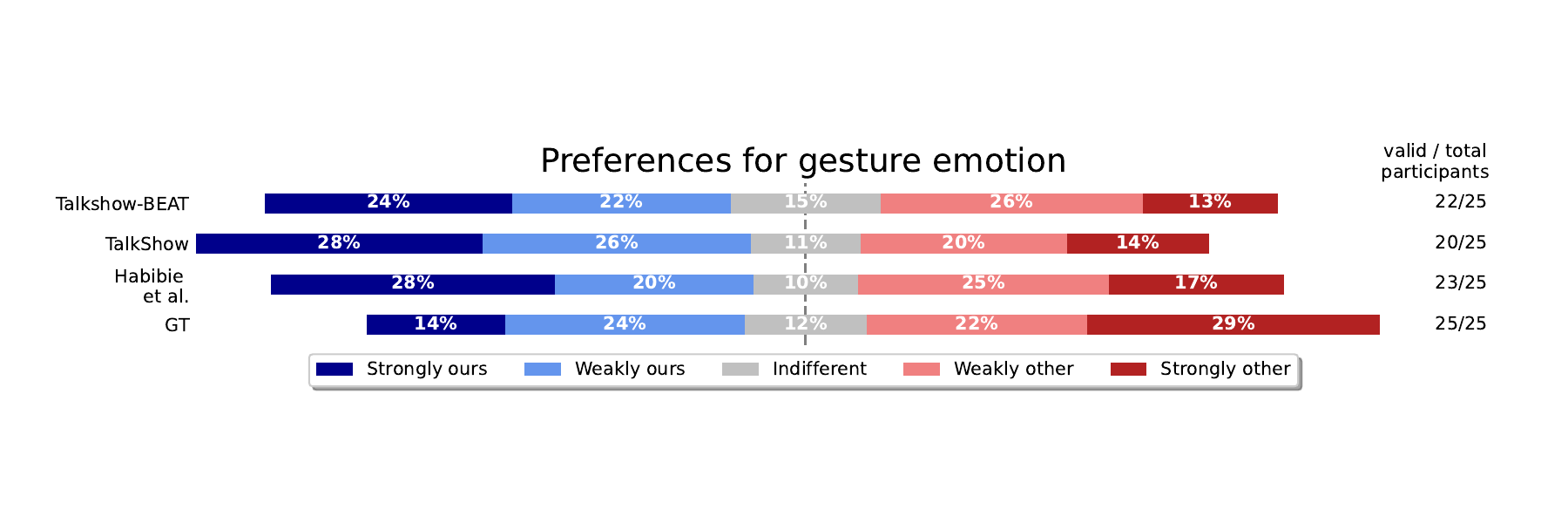}
   \includegraphics[trim={3.7cm 2.5cm 1.5cm 2.5cm},clip, scale=0.35]{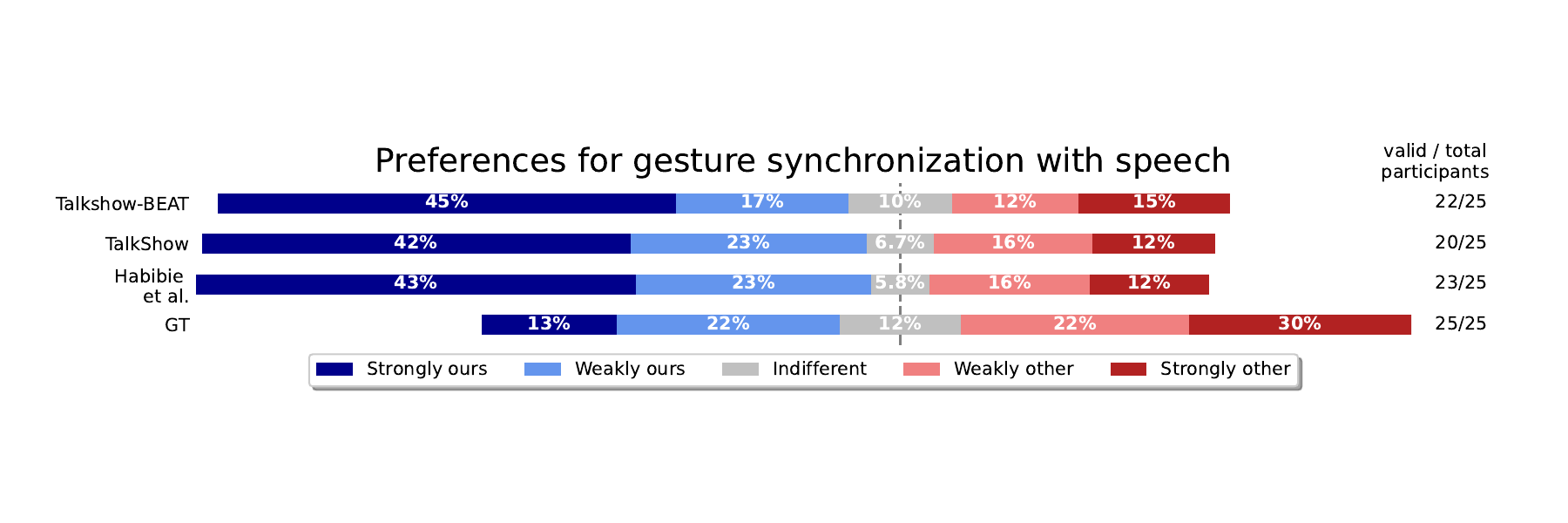}
   \vspace{-0.1in}
   \caption{
   \textbf{The perceptual study results for gesture emotion preference (left) and synchronization with speech (right)}. 
   The number of attentive participants that passed the catch trials is indicated on the right and the reported results only consider these participants.}
   \label{fig:perceptual}
\end{figure*}

\begin{figure}[t]
\centering
  \includegraphics[width=1\linewidth]{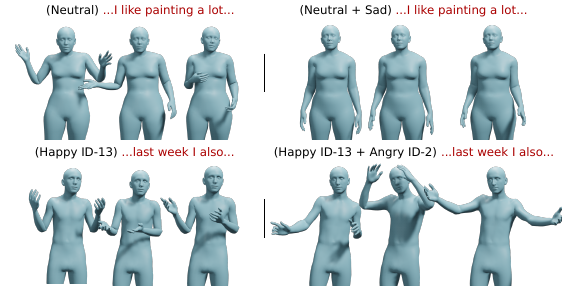}
   \caption{\textbf{Gesture editing.} Top:  We modify style from being neutral (left) to being sad (right) by combining the emotion latent from sad audio with the content latent from neutral audio. 
   Bottom: We transform the style from Subject 13 being happy (left) to being angry (right) by merging the content latent from happy audio with the style and emotion latents from an angry audio of Subject 2. 
    }
    \label{fig:ST}
\end{figure}

\paragraph{Emotion editing.}
We use two audio streams of a female subject for neutral and sad emotion. 
This experiment edits the subject's gesture style from moderately controlled hand movements to a sad style with lethargic posture conveying a sense of heaviness, as seen in~\cref{fig:ST} (top).

\paragraph{Gesture style editing.} 
We use audio streams of two male subjects for the happy (ID -  13) and angry (ID - 2) emotion. 
With the emotion, style and content latent fusion mechanism from two driving audio streams, \modelName is able to adapt the male (ID - 13) subject's body gestures from being close to their body to more open with squared tightened shoulders, expressing a shift from happy to angry emotions of a different subject (ID - 2), as shown in~\cref{fig:ST} (bottom). Please refer to the supplemental video for qualitative results and comparisons to additional gesture generation methods~\cite{liu2022beat,alexanderson2020style, Zhu_2023_CVPR} trained on coarse skeletal data. 

\subsection{Perceptual Study}
\label{subsec:perceptual}

\paragraph{Design.}
Our perceptual study is designed as a side-by-side comparison of two gesture videos generated with the same audio as input but by two different methods (\modelName and another model or GT). 
The participants are asked to rate their preference of the methods on a five-point Likert scale for ``synchronization with speech'' and ``gesture emotion appropriateness'' given the GT emotion label of the input audio.
We recruit 25 participants per method-to-method comparison on Amazon Mechnical Turk. 
Each participant is shown 24 pairs of randomly selected test set animations, 3 per emotion (neutral, happy, angry, sad, disgust, fear, surprise, and contempt). 
To allow the participant to get used to the task, we discard the answers of the first three comparisons and repeat these at the end. 
We incorporate three catch trials and responses from participants that fail on more than one are filtered out, as shown in Fig.~\ref{fig:perceptual} (right).

\paragraph{Results.}
The results of the study are shown in Fig.~\ref{fig:perceptual}. \modelName outperforms all competing methods by a considerable margin on both tasks, suggesting that \modelName's generations are more appropriate for both the content of the input speech and its emotion compared to the baselines. 
However, it must be noted that there is still a significant gap between \modelName and the GT.
Please refer to the \supmat~for details about the perceptual study.

\subsection{Discussion and Future Work}

\paragraph{Upper-body motion.}
\rebut{We focus on the 
smooth coordination between the pelvis and upper body animation for side-by-side comparisons with other methods, as all other methods primarily focus on upper body movements.}
Future work should include lower-body motion 
and locomotion 
as these impact the perceived emotional state of a sequence.

\paragraph{Semantics.}
While the generated gestures, synchronized with the driving speech sequence, do not account for semantics such as deictic 
and metaphoric gestures,
\rebut{incorporating the text/language modality could help further improve in this direction.}

\paragraph{Facial expressions.}
While emotional speech-driven face animation methods \cite{danecek2023emotional, peng2023emotalk} can be combined with bodies generated from \modelName, jointly learning to generate emotional 3D bodies from speech is a topic that needs attention.

\rebut{\paragraph{End-to-end training.}
Joint audio-gesture training may enhance results but requires careful loss term balancing and increased GPU memory. Therefore, we opted for separate training.}

\section{Conclusion}
\label{sec:conclusion}
We present \modelName, a framework to generate emotional body gestures from speech.
The emotions and personal styles of the synthesized gestures can be controlled, thanks to the disentanglement of content, emotion, and style directly from the speech. 
The latent diffusion-based framework can further generate variations of the same gesture with the same emotion.
Our quantitative evaluations show that \modelName achieves state of the art performance on a variety of metrics: diversity, gesture emotion classification accuracy,  Frechét gesture distance, beat alignment score, and semantic relevant gesture recall.
Finally, our perceptual study demonstrates that \modelName generates motions that are better synchronized and better match the emotion expressed of the input speech than previous state of the art.

\bigskip
\enlargethispage{2\baselineskip}

\begingroup
\small
\noindent\textbf{Acknowledgments.}
We thank A Cseke, T McConnell and T Alexiadis for help with the perceptual study, B Pellkofer, J Piles-Contreras, and E Fritzler for IT support, and P Kulits, M Petrovich, S Kokane, and S Zojaji for proofreading and insightful discussions.
This work was supported by the European Union’s Horizon 2020 research and innovation programme under the Marie Skłodowska-Curie grant agreement No.\ 860768 and by the Swedish Research Council through grant 2020-05187.

\noindent\textbf{Disclosure.}
\href{https://download.is.tue.mpg.de/amuse/Disclosure.txt}{https://download.is.tue.mpg.de/amuse/Disclosure.txt}
\par
\endgroup

{
    \small
    \bibliographystyle{ieeenat_fullname}
    \bibliography{references}
}

\bigskip
{\noindent \large \bf {APPENDIX}}\\
\renewcommand{\thefigure}{A.\arabic{figure}} 
\setcounter{figure}{0} 
\renewcommand{\thetable}{A.\arabic{table}}
\setcounter{table}{0} 

\appendix

This supplementary material summarizes the video content in~\cref{sec:supmatvid} and provides additional technical details of the speech disentangled model and the gesture generation model in~\cref{sec:audioX} and~\cref{sec:gestureX}, respectively. 
We provide details about motion extractor model in~\cref{sec:extractX}, discussions on the gesture emotion and semantics in~\cref{sec:delimitationX}, details on the data preparation process in~\cref{sec:dataprepX}, a review of state of the art methods in~\cref{sec:sotaX}, and additional information about the perceptual study in~\cref{sec:percepX}.
\section{Supplementary Video} 
\label{sec:supmatvid}

The supplementary video shows the generated gestures.
Specifically, it provides:
\begin{enumerate}
\item Gesture generations on various emotional audios,
  \item Gesture emotion and style editing results, 
  \item Comparisons with state of the art mesh-based and skeleton-based gesture generation methods,
  \item Ablation comparisons of the different components of our approach, 
  \item Gestures showing the diversity in the generations, and
  \item Gestures generated from an in-the-wild audio sequence.
\end{enumerate}

\section{Speech Disentanglement Model} 
\label{sec:audioX}

We explain the overall architecture in~\cref{subsec:astarchsec} and the encoder--transformer architecture in~\cref{subsec:dietencs}. We demonstrate the reconstruction mechanism to enforce disentanglement in~\cref{subsec:reconsconcatsec}. Finally, we explain the training procedure and loss terms in~\cref{subsec:asttrainsec}.

\subsection{Architecture}
\label{subsec:astarchsec}

\begin{figure}[t]
\centering
  \includegraphics[width=1\linewidth]{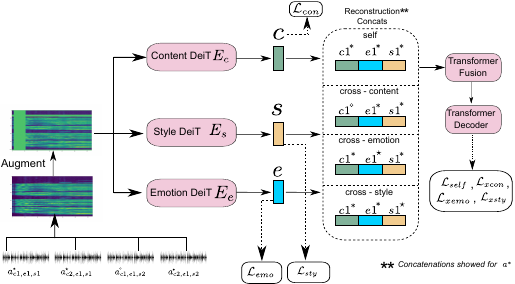}

   \caption{\textbf{Speech disentanglement model}. An input filterbank is given to the three encoders, producing three disentangled latents, which are decoded into a reconstructed filterbank. We here show disentanglement reconstruction for one audio only, please refer~\cref{subsec:reconsconcatsec} for its detailed explanation.}

 \label{fig:audionetarch}
\end{figure}

We illustrate speech disentanglement model architecture in~\cref{fig:audionetarch}. The training is conducted over audio of the same utterances spoken under different emotions or spoken by different speakers. Our model consists of three transformer encoders, a transformer fusion, and a transformer decoder. The input filterbank is simultaneously passed through content $E_c$, style $E_s$, and emotion $E_e$ transformer encoders, producing three disentangled latents: content $c$, style $s$, and emotion $e$.  The fusion and decoder are transformer-based layers. The transformer--fusion creates a single embedding by applying cross attention on the input triplet embeddings $( c,e,s )$. Finally, the transformer--decoder  reconstructs the original filter bank from the compressed single latent embedding produced by the transformer fusion. 

\subsection{Encoder Transformers}
\label{subsec:dietencs}

\begin{figure}[t]
 \centering
 \includegraphics[width=1\linewidth]{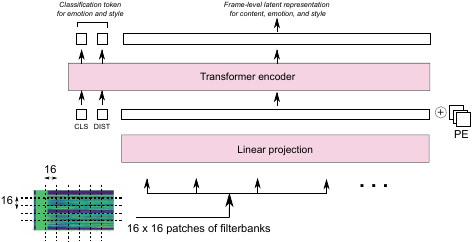}
   \caption{\textbf{Speech encoder transformer}. We have used encoder architecture based on Touvron et al.~\cite{TouvronCDMSJ21deit}. We use this architecture as content, emotion, and style encoders in the speech disentanglement model. Following Gong et al.~\cite{gong2022ssast, gong2021ast}, we use 10s augmented speech filterbank and split into fixed 1209 patches of 16 x 16 each, having 6 units overlap in frequency and time domain. The filterbank is passed through a linear projection layer and a learnable positional embedding (PE) is added to it.}
    \label{fig:astarch}
\end{figure}

Similar to~\cite{gong2022ssast, gong2021ast, palanisamy2020rethinking,guzhov2020esresnet,imageinmusic,Gong_2021} we employ transfer learning of vision task to our audio task by using pretrained weights of DeiT~\cite{TouvronCDMSJ21deit} (88M params) transformer that is fine-tuned on 384x384 images from ImageNet-1k~\cite{imagenet15russakovsky}. 
We use a pretrained DeiT encoder as a component of each of the encoders, as illustrated in~\cref{fig:astarch}. We linearly embed patches to features embedding of size 768 and feed them into DeiT along with trainable positional embedding of same size (768). 
We append class token $\textup{[CLS]}$ and distillation token $\textup{[DIST]}$ obtained from DeiT at the beginning of each filter bank sequence. 
We then average the 3-channel inputs of DeiT to obtain a single filterbank channel input. Finally, we use the output of the last DeiT encoder layer and project to 1D latent vector of 256 dimensions each, as our content, emotion, and style latents. 
We average the $\textup{[CLS]}$ and $\textup{[DIST]}$ tokens from DeiT and use it for audio emotion as well as audio style classification tasks for 8 and 30 category labels respectively.

\subsection{Reconstruction Concatenations}
\label{subsec:reconsconcatsec}
\cref{fig:concat} demonstrates a detailed information of the cross-reconstruction mechanism to enforce the audio content, emotion, and style disentanglement. Each audio in the quadruple is encoded and decoded to produce the reconstructed audio filterbank. To enforce content disentanglement, we swap the content latent vectors between different-subject same-emotion audio pairs with same utterances. Similarly, we swap emotion and style latents using audio pairs from the same subject. Specifically, we swap emotion latent vector and style latent vector between same-subject same-emotion audio pairs with different utterances. The procedure is repeated for each audio in the audio quadruples.

\subsection{Training and Losses}
\label{subsec:asttrainsec}
\begin{figure}[t]
\centering
\includegraphics[width=1\linewidth]{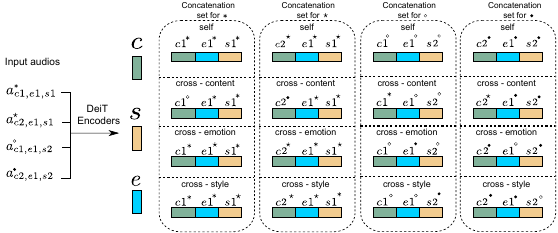}
   \caption{\textbf{Reconstruction concatenations for training forward pass}. We obtain disentangled content, emotion, and style latents from the transformer encoders. (\textit{Self}) concatenation of triplet latent vectors is used to decode back into the original filterbank. To enforce the content disentanglement, we swap content latent vectors (\textit{cross-content}) between given different-subjects audio pair with same utterances. Whereas to enforce style and emotion disentanglement, we swap style (\textit{cross-style}) and emotion (\textit{cross-emotion}) latent vectors between given same-subject audio pairs with same emotion categorical label. We repeat the procedure for quadruples of audio $\{ a^{\ast},a^{\star},a^{\circ},a^{\bullet} \}$ input in each forward pass.} 
    \label{fig:concat}
\end{figure}

We train the speech disentanglement model on 10s-audio segments of the BEAT dataset, which provides the GT labels for emotion and subject categorical labels. We split the audio data across actors during train, validation, and test step. 
During training, one sample is formed by a quadruple of
different audios $(a_1=a_{c_1,e_1,s_1}, a_2=a_{c_2,e_1,s_1}, a_3=a_{c_1,e_1,s_2}, a_4=a_{c_2,e_1,s_2})$, with two different contents $ c_1, c_2 $ 
(i.e., two different scripts),
two different styles $ s_1, s_2 $ 
(spoken by two different subjects)
and the same emotion $ e_1 $. 
To ensure, content, style, and emotion disentanglement, we employ a multitude of training losses. The self-reconstruction loss $\lossselfrec$ ensures that the style, emotion, and content latents extracted from the same audio can be decoded into the original inputs:

\begin{equation*}
     \lossselfrec=\sum_{k=1}^{4} \left \| \dec(\enccontent(a_{k}),\encstyle(a_{k}),\encemotion(a_{k}))  - a_{k} \right \|_{1}    
  \label{eq:rec-self}
\end{equation*}

The content loss $\losscontent$ ensures that two content latents extracted from two different audios with the same content $c_k$ but two different styles $ s_i, s_j $ match:

\begin{equation*}
 \losscontent=\sum_{k=1}^{2}\left \| \enccontent(a_{k})  - \enccontent(a_{k+2}) \right \|_{1}  
  \end{equation*}

We also employ the emotion classification loss $\lossemotion$ to ensure that the encoded emotion latents carry the emotion information. 
This is ensured by projecting them with a linear projection head into a classification vector and then computing emotion classification cross entropy loss. 
We use the same procedure to employ the style classification loss $\lossstyle$:

\begin{equation*}
\!
\begin{aligned}
   \lossemotion &= -\mkern-18mu\sum_{\substack{1 \leq l_e \leq n_e}}\mkern-18mu y_{l_e}\log(p_{l_e}),\\
  \mkern9mu \lossstyle &=-\mkern-18mu\sum_{\substack{1 \leq l_s \leq n_s}}\mkern-18mu y_{l_s}\log(p_{l_s}),
\end{aligned}
\label{eq:emo-sty-cla}
\end{equation*}

with $n_e=8$ and $n_s=30$ denoting the number of emotion classes and training subjects respectively.

Finally, we employ the cross-reconstruction losses for emotion, style, and content. 
This loss ensures that we can combine any three style, content, and emotion latents and decode them into  a valid reconstruction. 
As shown in~\cref{fig:audionetarch} and~\cref{fig:concat}, this is a three part cross reconstruction process. 
In this process, we extract content $\enccontent(a_{*})$, emotion $\encemotion(a_{*})$, and style $\encstyle(a_{*})$ latents of all four different audios. 
Given two input audios of the different contents $c_i$ and $c_j$, with the same speaker, and the same emotion, we swap the emotion latents between the audio pair, and decode the two audios back. 
Since the emotion class is constant within a quadruple, the emotion cross-reconstruction should be equal to the original audio. Similarly, we cross-reconstruct an input audio with two style latents of the same person, but of different sequence. Enforced by:

\begin{equation*}
\begin{aligned}
\losscrossemo &= \sum_{k=1}^{4} \dec(\enccontent(a_{k}), \encstyle(a_{k}), \encemotion(a_{j(k)})) - a_{k},\\
\losscrosssty &= \sum_{k=1}^{4} \dec(\enccontent(a_{k}),\encstyle(a_{j(k)})),\encemotion(a_{k}))   - a_{k},
\end{aligned}
\label{equ:rec-cross-emo}
\end{equation*}

where $j(k) = [(6 - k) \mod 4] + 1$. 

Given two input audios of the same contents, different speakers $s_i$ and $s_j$, and same emotion, we swap the content latents between the audio pair, and decode the two audios back. Since the utterances being spoken are the same and we keep the original style and emotion constant, the cross reconstruction for the swapped content should be equal to original audio. This is enforced by:

\begin{equation*}
\losscrosscon = \sum_{k=1}^{4}\dec(\enccontent(a_{j(k)})),\encstyle(a_{k}),\encemotion(a_{k}))   - a_{k} 
  \end{equation*}
where $j(k) = [(1 + k) \mod 4] + 1$. 

The combined audio loss is given as:

\begin{equation*}
\begin{split}
\mathcal{L}_{dis} & =\losscrosscon + \losscrossemo+ \losscrosssty \\
& +\lossselfrec+\lossemotion+ \losscontent+\lossstyle
\end{split}
  \label{eq:audioloss}
\end{equation*}

Once trained, the speech disentanglement model produces three disentangled latents for content, style and emotion. These latents serve as the input to our diffusion model.

\subsection{Implementation Details}

The encoder transformer DeiT (88M parameters) that is finetuned on 384x384 images from ImageNet-1k is obtained from PyTorch image models (timm)~\cite{fastaitimm}. The content, emotion, and style latent vectors are of 256 dimension. The transformer--fusion includes 2 layers and 4 heads. The  transformer--decoder includes 4 heads and 4 layers. The input dimension of fusion block is 768 to accommodate three content, emotion, and style latent codes. Each 2D filterbank is of 1024 x 128, where 128 represents the number of mel-frequency bins.

\begin{table}[b]
\caption{\textbf{\rebut{Audio emotion and style disentanglement ablation.}} We show scores for Emotion Accuracy (EA), Style Accuracy (SA), Emotion F1 Score (EF1), and Style F1 Score (SF1) in our speech disentanglement model and ablation experiments. Although there are slight differences, our model effectively captures the complex relationships between emotion and style by disentangling three latent vectors simultaneously. The best scores are highlighted in \colorbox{green!25}{green} and second best in \colorbox{blue!25}{blue}.}
\centering
\scalebox{0.85}{
\begin{threeparttable}
\begin{tabular}{lcccc}\centering
Method & EA (\%)\;$\uparrow$ & EF1$\uparrow$ & SA (\%)\;$\uparrow$ & SF1$\uparrow$ \\
\midrule
    Ours & \cellcolor{blue!25}91.531& \cellcolor{blue!25}0.914& \cellcolor{blue!25}96.060 & \cellcolor{blue!25}0.960 \\
    Emo-disentangle& \cellcolor{green!25}91.966 & \cellcolor{green!25}0.918 & --- & --- \\
    Sty-disentangle & ---& --- & \cellcolor{green!25}96.095 & \cellcolor{green!25}0.961 \\
\bottomrule
\end{tabular}
\end{threeparttable}
}
 \label{tab:audioquantX}
\end{table}

\begin{table}[b]
\caption{\rebut{\textbf{Audio latent component factorization ablation.} Self and cross-reconstruction errors show comparable performance, suggesting that individual latents from different audio sources can be effectively combined to yield valid outputs.} The best scores are highlighted in \colorbox{green!25}{green} and second best in \colorbox{blue!25}{blue}.}
\centering
\scalebox{0.85}{
\begin{threeparttable}
\begin{tabular}{lcccc}\centering
Method & Self$\downarrow$ & XCon$\downarrow$ & XEmo$\downarrow$ & XSty$\downarrow$ \\
\midrule
    Ours & \cellcolor{green!25}.3739& \cellcolor{green!25}.3740& \cellcolor{green!25}.3816 & \cellcolor{green!25}.3815 \\
    Emo-disentangle& .3793  & .3792  & \cellcolor{blue!25}.3905 & --- \\
    Sty-disentangle & \cellcolor{blue!25}.3769& \cellcolor{blue!25}.3770 & --- & \cellcolor{blue!25}.3887 \\
\bottomrule
\end{tabular}
\end{threeparttable}
}
 \label{tab:audioReconErr}
\end{table}

\subsection{Ablation Experiments}

\begin{figure*}[t!]
    \centering
    \captionsetup{type=figure}
    \includegraphics[width=1\textwidth]
    {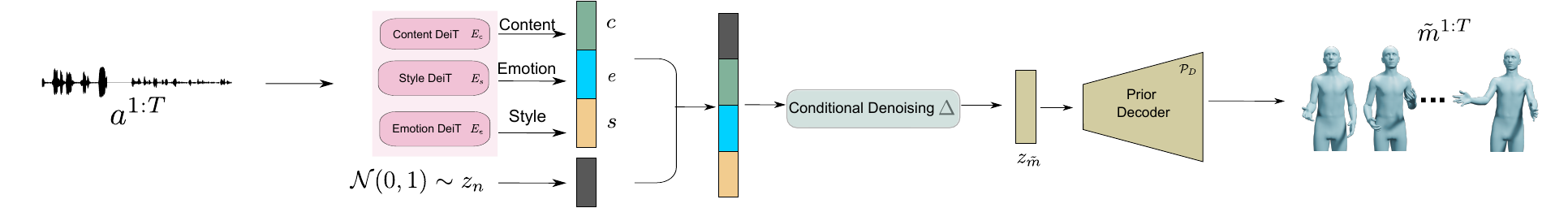}
    \captionof{figure}{
    \textbf{Inference.} We sample $\normalvec$ and employ the three conditioning latents from a test-time audio $\tmplatentcont, \tmplatentemo, \tmplatentstyle$. We iteratively apply $\denoisernet$ to generate the fully denoised $\denoisedlat$ which is decoded by $\priordec$ into the final motion $\diffdecodemotion$.
    }
    \label{fig:Amuseinfer}
\end{figure*}
\begin{figure}[t]

  \centering
  \includegraphics[width=0.85\linewidth]{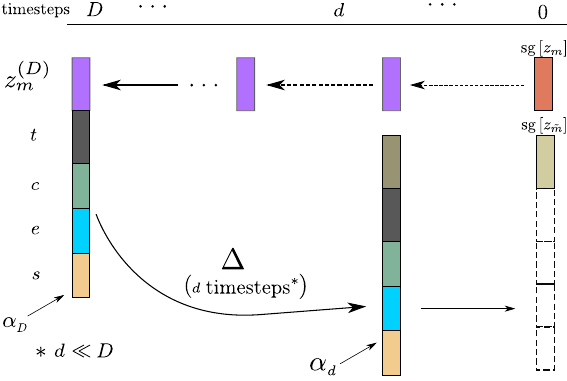}

   \caption{\textbf{Conditional latent diffusion.} In the diffusion process (right to left) we obtain a noisy motion latent, whereas in the denoising process (left to right) we obtain a conditioned denoised motion latent.}
   \label{fig:ddpimx}
\end{figure}

We conduct two ablation studies with the speech disentanglement model.
\rebut{One to only disentangle emotion from content (dropping $\mathcal{L}_{sty}$, $\mathcal{L}_{xsty}$). 
The other to only disentangle only style from content (dropping $\mathcal{L}_{emo}$, $\mathcal{L}_{xemo}$).}~\cref{tab:audioquantX} shows the accuracy and F1 scores for emotion and style latent vectors.  The Emotion Accuracy (EA), Style Accuracy (SA), Emotion F1 Score (EF1), and Style F1 Score (SF1) in our speech disentanglement model exhibit only marginal differences compared to the results obtained in the ablation experiments. \rebut{We report the test set self- and cross-reconstruction errors in~\cref{tab:audioReconErr}. 
The cross-reconstruction errors are comparable to self-reconstruction errors which indicates that the individual latents from different audios can be combined to produce valid outputs.
This holds for the main model and also the ablated models. 
However, the ablated models are not able to factor the audio into all three components due to the dropped loss terms.}We observe the robust performance of our audio model, by accounting for the complex interplay between emotion and style. By concurrently disentangling three latent vectors, our approach effectively captures the intricate relationships in the audio data, allowing to jointly model and distinguish both emotion and style factors.

\section{Gesture Generation Model}
\label{sec:gestureX}

\subsection{Motion Prior And Latent Denoiser}
\begin{figure}[t]
  \centering
   \includegraphics[width=1\linewidth]{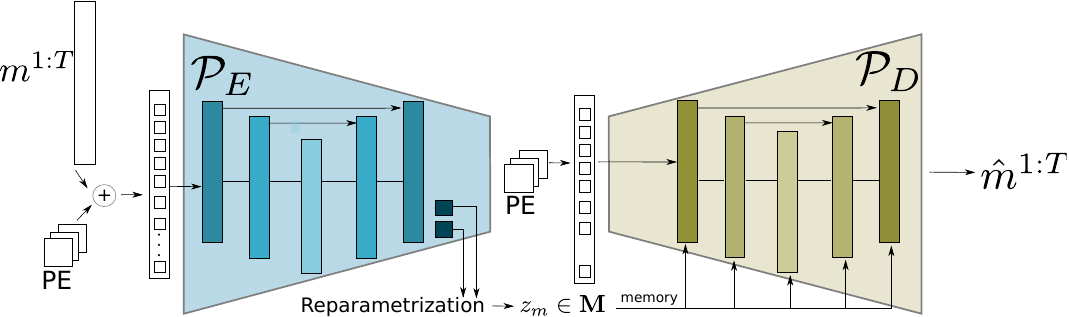}
\caption{\textbf{Motion prior network}. The motion prior is VAE encoder decoder architecture inspired from Chen et al.~\cite{chen2023executing}. Both encoder and decoder follow a U-Net like structure with skip connections between transformer blocks. The learnable positional embeddings (PE) are injected into each multi-head attention layer.
   }
   \label{fig:priornetX}
\end{figure}

In this section we include detailed illustrations of the motion prior and latent denoiser.~\cref{fig:Amuseinfer} illustrates the inference process employed by our model.~\cref{fig:ddpimx} illustrates the forward diffusion and the reverse audio-conditioned denoising process, operating at the latent space. Finally,~\ref{fig:priornetX} shows the diagram of the architecture of the motion prior.

\begin{figure}[t]
  \centering
   \includegraphics[width=0.85\linewidth]{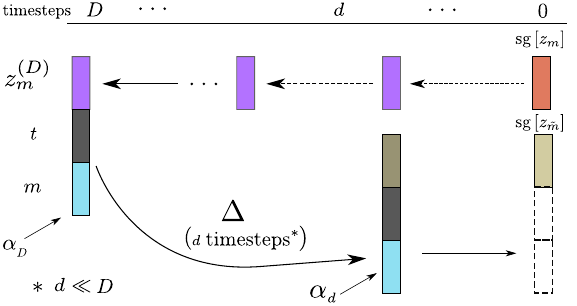}
   \caption{\textbf{Ablation conditional latent diffusion.} In the denoising process (left to right) of the ablation model, we obtain a denoised motion latent  that is conditioned on a compressed non-disentangled latent vector $m$ instead of three disentangled latents that are used in the final model.}
   \label{fig:wodisablation}
\end{figure}

\subsection{Methods Trained on Coarse Skeletal Data}

\begin{figure}[t]
  \centering
   \includegraphics[width=0.6\linewidth]{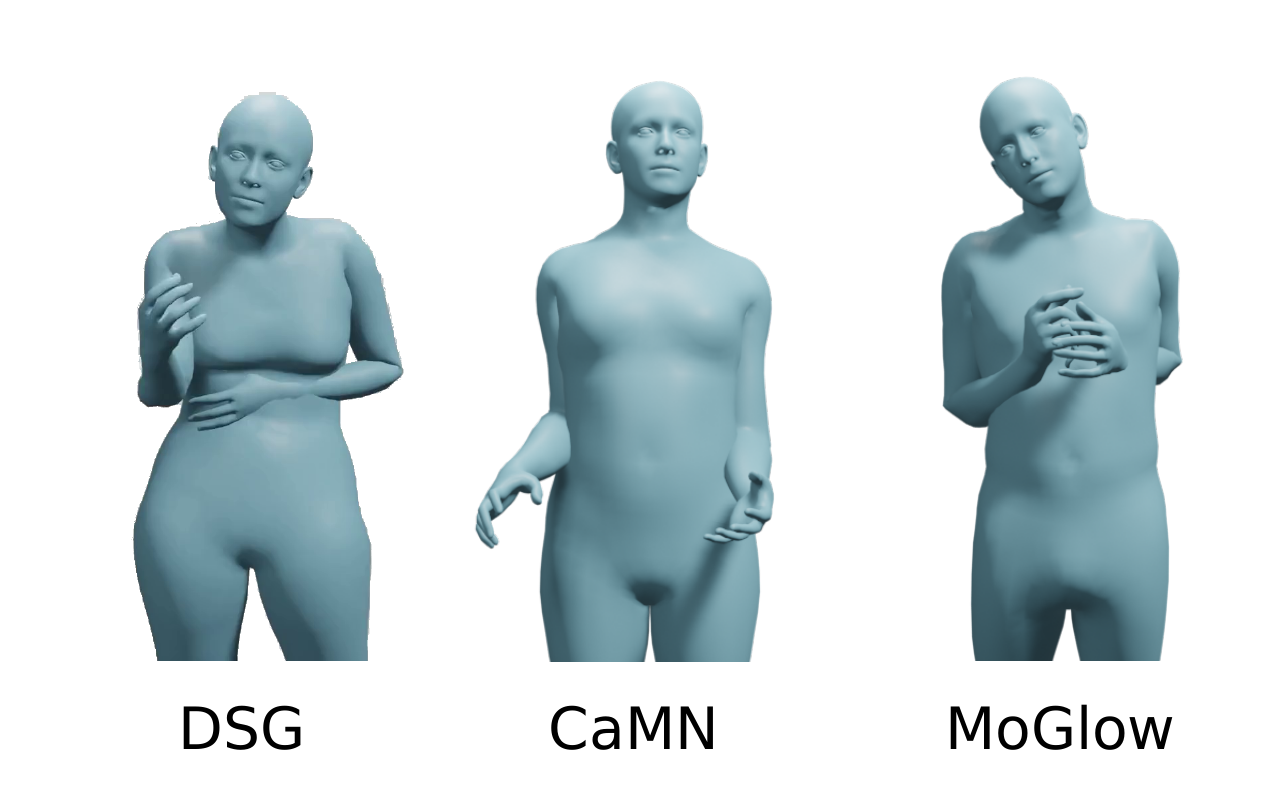}

   \caption{\textbf{Coarse skeleton-based methods.} 
   Here we compare DSG~\cite{yang2023DiffuseStyleGesture}, CaMN~\cite{liu2022beat}, and MoGlow~\cite{alexanderson2020style}. 
    Unlike SMPL-X-based models, these are trained using different skeletal hierarchies without volumetric 3D shapes. 
    Retargeting them onto the SMPL-X skeleton with IK causes uncanny poses and self-penetration. 
   }
   \label{fig:camn}
\end{figure}

We compare \modelName with methods trained on coarse skeletal data. 
We choose DSG~\cite{yang2023DiffuseStyleGesture},  CaMN~\cite{liu2022beat}, Zhu et al.~\cite{Zhu_2023_CVPR} and MoGlow~\cite{alexanderson2020style} as recent gesture generation models using audio input. \modelName produces more synchronized gestures and better represents the underlying audio emotion compared to the state of the art methods trained on skeletal data, as shown in our supplementary video. Additionally, these methods are not trained to output 3D meshes. We observe uncanny poses and self-penetrations as shown in~\cref{fig:camn}. In our video, we provide additional comparison with these skeleton based methods in both formats, the original predictions of those models and 3D meshes which are created via Inverse Kinematics (IK). We exclude the conversion to 3D mesh for Zhu et al.~\cite{Zhu_2023_CVPR} because the output skeleton format is incompatible with SMPL-X topology.

\subsection{Ablation Experiments}

\paragraph{Without speech disentanglement model}. This subsection illustrates the difference between the final \modelName model and the ablation of \modelName w/o audio disentanglement. \modelName w/o audio disentanglement uses 8 linear-layered auto-encoder that operates directly on raw audio MFCC features  to produce single latent vector $m$.
Since \modelName w/o audio disentanglement does not operate over the three disentangled latents of content, emotion, and style but instead only one non-disentangled latent $m$, the latent diffusion process also only takes one latent on  the input $m$ as shown in Fig.~\ref{fig:wodisablation}. By design, this model lacks the gesture editing capabilities.

\paragraph{Without motion prior}. We employ our latent denoiser only in this ablation model. We completely removed the motion prior component and replace it with a linear projection head.
The ablation model without motion prior is not able to converge and produces mostly static motions (refer to the supplementary video). This signifies the importance of having a motion prior component in our \modelName architecture.

\begin{table}[b]
\caption{\textbf{Ablation of \modelName components.} The model without audio disentanglement produces lower-quality gestures and lacks editing capabilities. The model without motion prior perform poorly due to convergence issues. Among the methods being compared, we highlight the best scores in \colorbox{green!25}{green} and second best in \colorbox{blue!25}{blue}. }
\centering
\scalebox{0.85}{
\begin{threeparttable}
\begin{tabular}{lccccc}\centering
    Method & SRGR$\uparrow$ & BA$\uparrow$  & FGD$\downarrow$  & Div$\rightarrow$  & GA\tnote{a}\;$\uparrow$  \\ %
\midrule
    GT  & --- & 0.83  & ---  & 27.83& 64.04  \\
\midrule
    Ours  & \cellcolor{green!25}0.36 & \cellcolor{green!25}0.81  & \cellcolor{green!25}388.63  & \cellcolor{green!25}25.06& \cellcolor{green!25}46.76  \\
    Ours-No-Prior & 0.25 & 0.20  & 987.90  & 13.41& 15.42 \\
    Ours-No-Audio-Model & \cellcolor{blue!25}0.31 & \cellcolor{blue!25}0.78  & \cellcolor{blue!25}633.27  & \cellcolor{blue!25}21.08& \cellcolor{blue!25}26.88 \\
\bottomrule
\end{tabular}
\begin{tablenotes}
            \item[a] GA is average of all 8 emotions.
        \end{tablenotes}
\end{threeparttable}
}
 \label{tab:quantgestablat}
\end{table}

\begin{table}[b]
\caption{\textbf{Ablation of \modelName training.} \rebut{We observe improved FGD and Div scores when the motion prior and diffusion model are jointly trained, highlighting superior performance compared to separate training methods.} We highlight the best scores in \colorbox{green!25}{green} and second best in \colorbox{blue!25}{blue}. }
\centering
\scalebox{0.85}{
\begin{threeparttable}
\begin{tabular}{lcc}\centering
    Method &  FGD$\downarrow$  & Div$\rightarrow$ \\ 
\midrule
    Ours  & \cellcolor{green!25}388.63 & \cellcolor{green!25}25.06   \\
    Ours-Disjoint & \cellcolor{blue!25}362.33 & \cellcolor{blue!25}24.49  \\
\bottomrule
\end{tabular}
\end{threeparttable}
}
 \label{tab:quantgestablatdisjoint}
\end{table}

\paragraph{Quantitative evaluation of ablation experiments}. Following the procedure described in the Sec. 5.1
, we report quantitative evaluation scores in~\cref{tab:quantgestablat}, comparing \modelName with the ablation models and GT. The version w/o speech disentanglement model produces lower-quality gestures and lacks editing capabilities compared to the complete model. This is because it lacks a component for separating emotion, content and style in the audio. The scores for the ablation models without motion prior are the lowest, indicating that this model did not converge successfully. \rebut{Additionally, in~\cref{tab:quantgestablatdisjoint} we report improved FGD and Div scores when the motion prior and diffusion model are trained jointly compared to when trained separately, indicating that joint training yields superior results. Furthermore, we conduct additional ablation experiments with and without alignment losses ($\mathcal{L}_{align}$, $\mathcal{L}_{Valign}$). Including alignment losses results in a GA of 46.79\%, whereas without them, the GA drops to 30.89\%, demonstrating the alignment losses effectiveness. Moreover, we compute the average jerk of the left and right hands for motion sequences belonging to the same audio of~\cite{yang2023DiffuseStyleGesture}, ours, and GT, reporting it in m/s\textsuperscript{3} as 1.18, 1.10, and 0.065, respectively. This signifies that the GT motion is the most steady, whereas ours is slightly smoother over time compared to~\cite{yang2023DiffuseStyleGesture}.}

\section{Motion Feature Extractor Model}
\label{sec:extractX}
We employ the motion extractor model $\mfeatxtractor$ for computing all quantitative evaluation metrics. Our motion extractor encoder model design is inspired by Petrovich et al.~\cite{petrovich21actor}, in an autoencoder setting (i.e., without a probabilistic variational component). We append a $\textup{CLS}$ token at the beginning of the motion sequence and supervise  with a cross-entropy emotion classification objective $\mathcal{L}_{Memo}$ applied to the output $\textup{CLS}$ token. We train the motion extractor model on the BEAT training data. Once trained, we use the latent space features to compute evaluation metrics as described in the Sec. 5.1.

\begin{equation*}
    \mathcal{L}_{Memo}= -\mkern-18mu\sum_{\substack{1 \leq l_e \leq n_e}}\mkern-18mu y_{l_e}\log(p_{l_e})
  \label{eq:extractorloss}
\end{equation*}

with $n_e=8$ denotes the number of emotion classes.

\section{Gesture Emotions And Semantics}
\label{sec:delimitationX}

We quantitatively evaluate our method using metrics SRGR, beat align, FGD, diversity, and gesture emotion accuracy. Leveraging the latent space features from the motion extractor model $\mfeatxtractor$, we compute SRGR and gesture emotion accuracy. Additionally, we directly utilize the generated motion sequence to calculate the beat align score.

\begin{figure*}[t]
    \centering
    \captionsetup{type=figure}
    \includegraphics[width=0.7\textwidth]{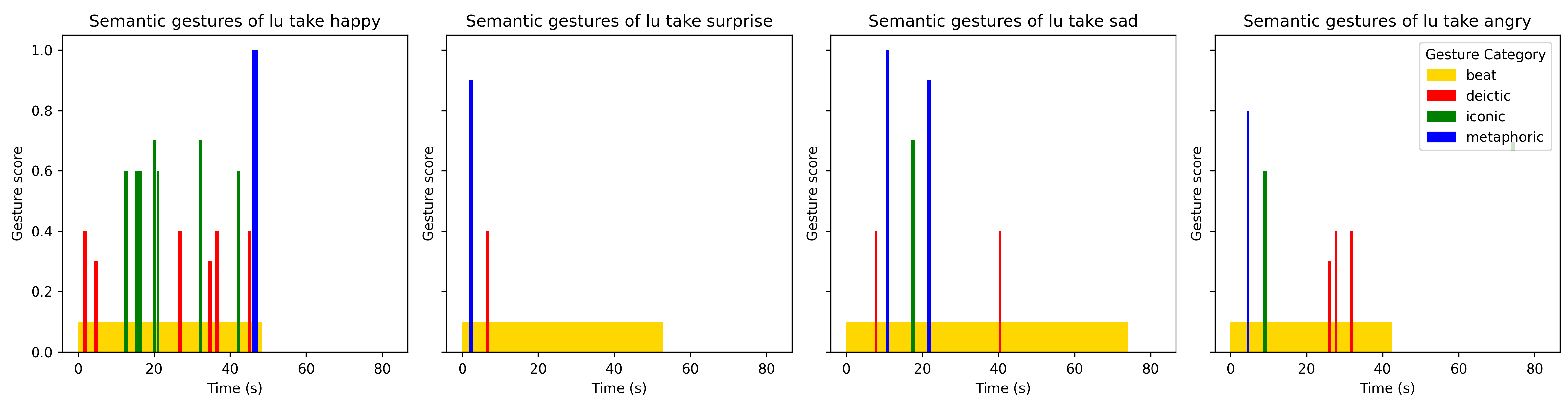}
    \captionof{figure}{\textbf{Emotional gesture variation.} Semantic scores for various emotions within the same subject shows how the subject expresses gestures differently for each emotion. This reflects the subject's interpersonal style specific to each emotion.
    }
    \label{fig:allemosem}
\end{figure*}
\begin{figure*}[t]
    \centering
    \captionsetup{type=figure}
    \includegraphics[width=0.85\textwidth]{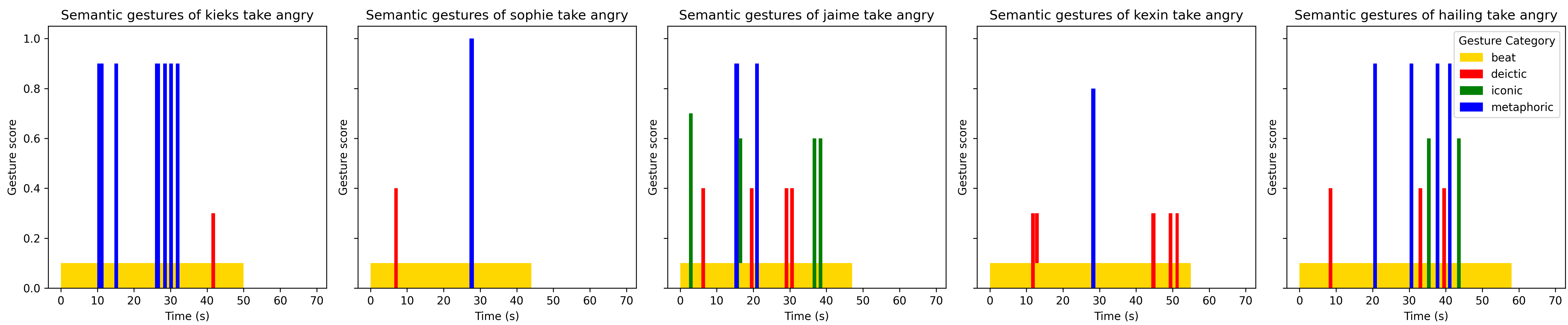}
    \captionof{figure}{\textbf{Emotional gesture individuality.} Semantic scores across various subjects for the same emotion reveal how different subjects express gestures uniquely for identical utterances within same emotion. There is variability in expressiveness, with some subjects being more expressive (eg. Jamie, Hailing) than the others (eg. Sophie, Kexin).
    }
    \label{fig:oneemoallsub}
\end{figure*}

\paragraph{Semantic-Relevant Gesture Recall (SRGR)}. In the SRGR metric score, similar to Liu et al.~\cite{liu2022beat}, we use ground truth semantic score as weight for the Probability of Correct Keypoint (PCK) between the generated gestures and ground truth gestures, where PCK is the number of joints successfully recalled for a given threshold $\delta$. Following the approach suggested by BEAT authors:

\begin{equation*}
    \textup{SRGR }= \lambda \sum \frac{1}{\textup{T x J}}\sum_{t=1}^{T}\sum_{j=1}^{J}\textbf{1}\left [ \left \| p_{t}^{j} - \hat{p}_{t}^{j}\right \|_{2} < \delta \right ]
  \label{eq:srgrmetric}
\end{equation*}
where \textbf{1} is the indicator function, $T$, $J$ are the set of frames and number of joints, we use SRGR to measure how well our model recalls gestures in the relevant clip. This metric reflects human perception of valid gesture diversity. The metric is computed based on the scores assigned by 118 annotators from Amazon Mechanical Turk (AMT), who evaluated the semantic relevance on a continuous scale of 0-1. The scores are provided for four gesture types: beat (\textit{rhythmic movements}), iconic (\textit{representative movements}), deictic (\textit{indicative or pointing movements}), and metaphoric (\textit{symbolic or figurative movements}). SRGR metric needs GT semantic scores for computation.

\paragraph{Ground-truth semantic scores}. We obtain the ground-truth semantic score, provided by the BEAT authors, for computing the SRGR. In \cref{fig:allemosem}, we present semantic scores for the same subject across various emotions, while \cref{fig:oneemoallsub} illustrates semantic scores for all subjects expressing the same emotion. This allows us to observe how subjects gesture differently with different emotions and how different subjects gesture for the same emotion. While we acknowledge the high-quality dataset introduced by the BEAT authors, our model has the potential to deliver even better results and improved expressivity with an enhanced dataset quality.

\paragraph{Beat alignment}. Following Li et al.~\cite{Li2021LearnTD}, we compute the beat align score. To compute the beat alignment score, we use six joints: left wrist, left elbow, left shoulder, right wrist, right elbow, and right shoulder, similar to Liu et al.~\cite{liu2022beat}. We measure the synchronization between the generated 3D motion and the input speech by calculating the beat align score. This score gauges the average distance between each kinematic beat and its nearest speech audio beat, following a unidirectional approach -- recognizing that gesture motion may not align with every speech audio beat. \modelName achieves the highest beat align score in correlating speech audio and gestures compared to the other methods.

\begin{figure}[t]
  \centering
   \includegraphics[width=1\linewidth]{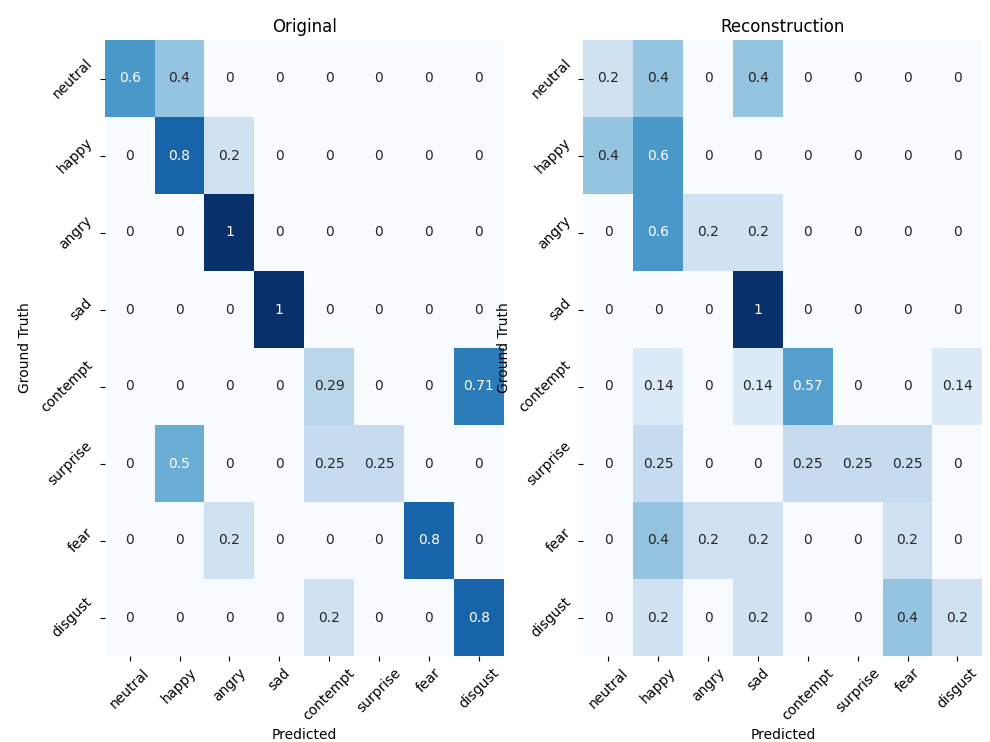}

   \caption{\textbf{Confusion matrix comparing gestures from the ground truth (GT) and regenerated emotion predictions}. }

   \label{fig:reconsCM}
\end{figure}

\begin{figure}[t]
  \centering
   \includegraphics[width=1\linewidth]{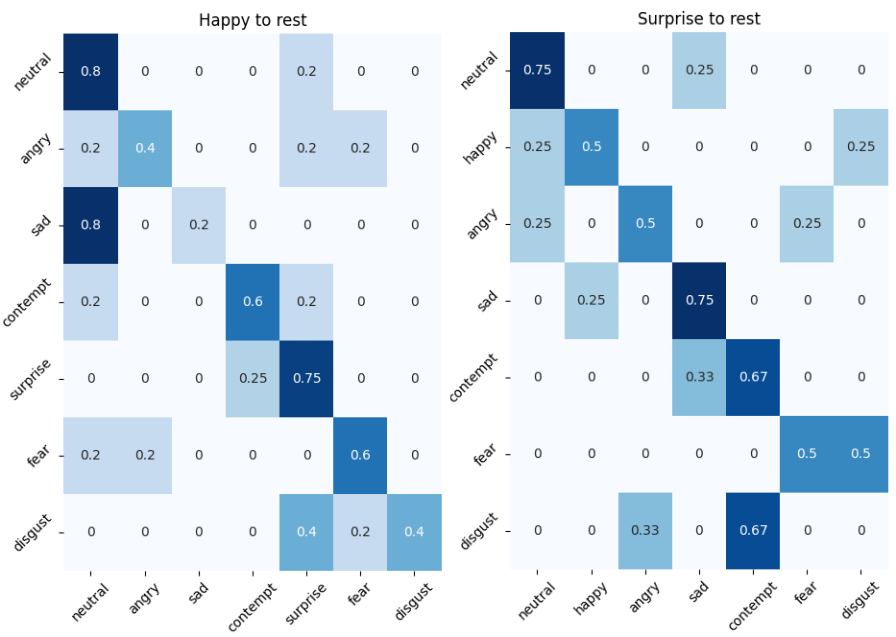}

   \caption{\textbf{Emotion edit confusion matrix, displaying transitions from Happy (left) and Surprise (right)  emotions to others}. X-axis is for predictions and Y-axis  is for ground truth.}
   \label{fig:editCM}
\end{figure}

\paragraph{Gesture emotion accuracy}. Gestural emotions are complex, influenced by internal states of subject, social signals, and their perception vary significantly across individuals with diverse cultural backgrounds. \modelName is designed to capture perceived gestural emotions. While we demonstrate \modelName with a gesture emotion recognition accuracy of 46.76\% and \modelName-Edit with 34.18\%, outperforming other state of the art methods, it is important to note that recognizing emotion from gestures remains a challenging task in computer vision. We observe the gesture emotion accuracy for the GT sequence is 64.04\%. There is still ample room for improvement in addressing this complex problem. Additionally, in~\cref{fig:reconsCM}, we present the confusion matrix for ground truth (GT) emotion predictions on the left and reconstructions (gestures generated using the original style, emotion, and content latents of a given audio) on the right. We observe a robust correlation between the predictions on GT and the reconstructions for all eight emotions. Additionally, we conducted experiments on gesture edits by swapping emotion latents from one audio with those from another audio of the same subject but with a different utterance. In~\cref{fig:editCM}, we showcase two exemplars, transforming from \red{\textit{Happy to rest}}  and \red{\textit{Surprise to rest}}. Given the diversity of eight emotions, gestural edits offer numerous possibilities, rendering this a broad and challenging problem. Although~\cref{fig:editCM} displays promising results for emotion label predictions with clear diagonal pattern  of the confusion matrix, we acknowledge the inherent difficulty in solving this problem.

\section{Data Preparation} 
\label{sec:dataprepX}

In this section, we describe the processing and alignment of different modalities, and the BEAT~\cite{liu2022beat} data subsets employed to train the different models of our framework. We do not use the entirety of the BEAT dataset to train \modelName. 
BEAT contains 30 speakers. 
We filter out subjects with little expressivity in their motion through visual inspection of GT, leaving us with 22 subjects.
Furthermore, BEAT has a subset that all the subjects speak the same sentences in the same emotions. 
The rest of the dataset contains unique sentences which are spoken only by one speaker and not the others. 
We filter out all of these unique sentences.
What remains is a subset of 16 sentences (2 per emotion, for 8 emotions), spoken by every subject. 
This is critical since the training of the speech disentanglement module requires perfect temporal correspondence between the audios of the same sentences. 
Except where explicitly stated otherwise, we have used this subset and split it into train, validation, and test sets.  
This subset is $5.71$ hours long.
We use the same data to train our speech disentanglement model. 
Further, we train our motion prior network (${\priorenc, \priordec}$) with the extracted SMPL-X motions of the same subsets. 
Finally, the denoiser, $\denoisernet$, and feature extractor used for evaluation, $\mfeatxtractor$, are trained on the same subset and splits.

\section{Review of State of the Art Methods}
\label{sec:sotaX}

\paragraph{Data\rebut{ selection and }input formats.}
\rebut{To train AMUSE effectively, we require data in the form of 3D point clouds rather than coarse BVH skeletons. Additionally, training requires common utterances from multiple subjects expressing various emotions for audio disentanglement. Many available gesture datasets, including~\cite{ghorbani2022zeroeggs,10.1145/3592456}, come in various motion capture skeleton formats with different underlying kinematic hierarchies that are incompatible with our conversion procedure to obtain SMPL-X meshes and do not meet the requirement of speech common utterances. In contrast, for the BEAT dataset, we obtained the initial data in the form of 3D point clouds from the dataset authors. We use Mosh++~\cite{AMASS:2019, Loper:SIGASIA:2014} to extract SMPL-X pose and shape parameters, along with global translation and orientation, from the 3D point cloud. This data was     then used to train AMUSE.}

\begin{table}[b]
\caption{\textbf{Perceptual study.} We demonstrate aggregate scores of our perceptual study. and we disregard indifferent scores. The \textit{ours} and \textit{others} are sum of \% preference for (strongly ours and weakly ours) and (strongly other and weakly other) respectively. Only the best scores are highlighted in \colorbox{green!25}{green}.}
\centering
\scalebox{0.8}{ 
\begin{threeparttable}
\begin{tabular}{llllll}\centering
\textbf{Criteria}\;$\rightarrow$ & \multicolumn{2}{l}{Emotion}   &|&\multicolumn{2}{l}{Synchronization}  \\
\textbf{Method}$\;\downarrow$ & Ours & Others  & |&Ours  &  \multicolumn{1}{c}{Others}  \\
\midrule
    GT & 38&\cellcolor{green!25}\;\;\;51& | & 35 & \cellcolor{green!25}\;\;\;\;\;\;\;52 \\
    TalkSHOW-BEAT & \cellcolor{green!25}46&\;\;\;39 & |  &\cellcolor{green!25}62 & \;\;\;\;\;\;\;27\\
    TalkSHOW & \cellcolor{green!25}54&\;\;\;34 &   |&\cellcolor{green!25}65 & \;\;\;\;\;\;\;28\\
    Habibie et al. & \cellcolor{green!25}48&\;\;\;42& |  &\cellcolor{green!25}66 & \;\;\;\;\;\;\;28 \\
\bottomrule
\end{tabular}
\end{threeparttable}
}
 \label{tab:AMTavg}
\end{table}

\paragraph{SOTA methods and modifications}.
Given our primary objective is to generate 3D emotional gestures from audio input, we mainly compare state-of-the-art methods that use audio input alone and output a 3D mesh. 
We exclude methods that incorporate additional inputs, such as arbitrary lengths of target motion style, as they deviate from our main objective, for example, Ghorbani et al.~\cite{ghorbani2022zeroeggs}. Other recent works~\cite{Ao2023GestureDiffuCLIP,alexanderson2023listen} have proposed methods for generating gestures from speech. However, making direct comparisons is difficult as the code for their approaches is not publicly available. We retrained Henter et al.~\cite{henter2020moglow} using publicly available code and instructions, due to the unavailability of a pretrained model. In our comparison, we used publicly available DSG~\cite{yang2023DiffuseStyleGesture} model that was trained on the BEAT dataset of coarse skeletal format. We also made modifications to the TalkSHOW code, incorporating emotion labels as input, and retrained it on the same data used for training our model. The emotion categorical labels were injected inline with existing subject labels using one-hot vectors. \modelName outperforms both DSG and TalkSHOW-BEAT as well as other SOTA methods in all comparisons.

\section{Additional Perceptual Study Details}
\label{sec:percepX}

Here we describe additional details of the AMT study reported in the Sec. 5.3.
We show aggregate preference scores in~\cref{tab:AMTavg}. \modelName outperforms all methods compared against in both criteria - synchronization with the speech and the appropriateness with respect to the specified emotion. In contrast, Ground Truth (GT) consistently outperforms \modelName in both tasks. This outcome emphasizes the complexity of the problem, where achieving synchrony with speech and meeting specified emotional appropriateness remain challenging objectives.

\paragraph{Data.}
We randomly select three videos per emotion from the BEAT dataset for our perceptual study. 
We only use sequence that were not part of training or validation set.
Due to high number of subjects, we limit the input audios data to only two subjects.

\begin{figure*}[t]
    \centering
    \captionsetup{type=figure}
    \includegraphics[width=1\textwidth]{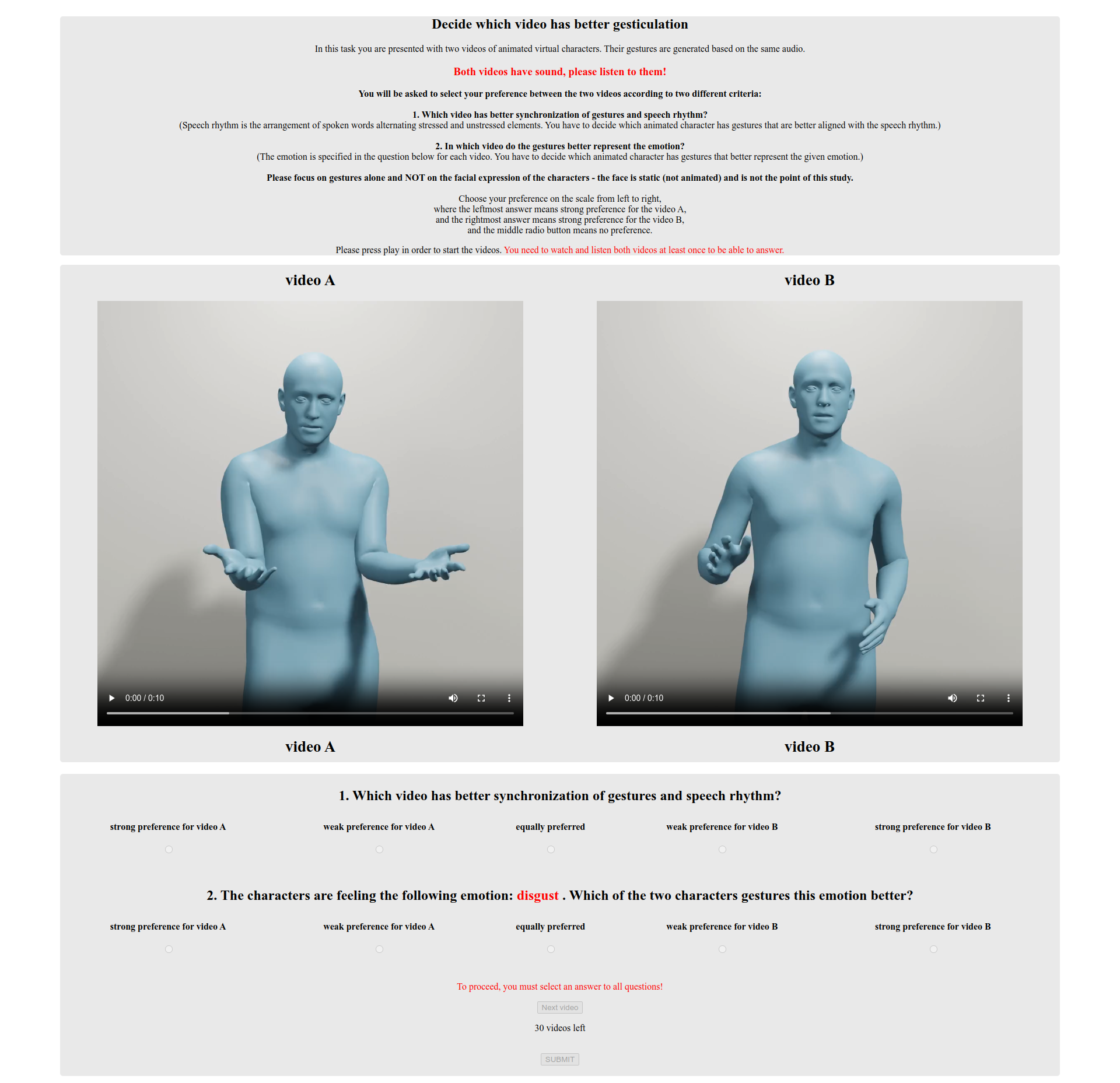}
    \captionof{figure}{\textbf{The layout of the perceptual study.} The participant is shown two videos and asked to enter their preference according to two criteria - synchronization with the speech and the appropriateness with respect to the specified emotion.
    }
    \label{fig:study_layout}
\end{figure*}

\paragraph{The template layout.} 
Fig.~\ref{fig:study_layout} depicts the design template that the participants were shown. 
The left--right position of our method and the competing methods was randomized to factor out any biases that participants may have for one side or the other. 

\paragraph{Catch trials.}
Each participant was also shown three catch trials, where a GT video was shown alongside a broken motion filled with artifacts.
Participants that did not select weak or strong preference for the GT video in any of the catch trials were labeled as uncooperative or inattentive and were not considered in the analysis. We selected 22, 20, 23, and 25 participants for TalkSHOW-BEAT, TalkSHOW, Habibie et al., and GT, respectively, from a total of 25 Amazon Mechanical Turk workers.

\end{document}